\PassOptionsToPackage{draft,bookmarks=false}{hyperref}
\documentclass[letterpaper, 10 pt, journal, twoside]{ieeetran} 
\IEEEoverridecommandlockouts                               
\usepackage{etoolbox}
\makeatletter
\patchcmd{\@makecaption}
  {\scshape}
  {}
  {}
  {}
\makeatother

\IEEEoverridecommandlockouts                               

\usepackage{bm}

\usepackage{graphicx}                       
\usepackage{graphics}                       
\usepackage{booktabs}
\usepackage{graphicx}
\usepackage{subcaption}
\usepackage{epsfig}                         
\graphicspath{./pics/RAL_fig_for_revise}
\graphicspath{./pics/RAL_new_fig}
\usepackage{enumitem}
\usepackage{amssymb,amsmath}
\usepackage{mdwmath}
\usepackage{commath}   
\usepackage{eqparbox}
\usepackage{mathtools}
\usepackage[utf8]{inputenc} 
\usepackage[english]{babel}
\usepackage{amsmath,amsfonts,amssymb}

\newcommand{\beq}{\begin{equation}}
\newcommand{\eeq}{\end{equation}}
\newcommand{\bear}{\begin{eqnarray}}
\newcommand{\bears}{\begin{eqnarray*}}
\newcommand{\eear}{\end{eqnarray}}
\newcommand{\eears}{\end{eqnarray*}}
\newcommand{\bdm}{\begin{displaymath}}
\newcommand{\edm}{\end{displaymath}}
\newcommand{\lba}{\left[\begin{array}}
\newcommand{\ear}{\end{array}\right]}

\usepackage{tabularray}

\usepackage{stfloats}                       
\usepackage{url} 
\usepackage{hyperref}
\usepackage{cite}                           
\usepackage[T1]{fontenc} 
\usepackage{tabularx}
\usepackage{diagbox} 
\usepackage{color}
\usepackage{multirow}
\usepackage[table]{xcolor}
\usepackage{longtable}
\usepackage{booktabs} 			
\usepackage{xcolor,colortbl}    

\usepackage{cuted}
\usepackage{capt-of}

\usepackage{multicol}
\usepackage{graphicx}
\usepackage{placeins}
\usepackage{overpic}
\usepackage{amssymb}
\usepackage{pifont}

\usepackage[fleqn]{nccmath}

\usepackage{amsmath}
\usepackage{algorithm}
\usepackage{algpseudocode}

\title{{\fontsize{23}{1}\selectfont TactileLab: Sim-to-Real Dexterous Robot Manipulation\\
with Efficient Shear-Sensitive Tactile Simulation}}

\author{%
Yijiong Lin$^{1,2}$,
Wenhao He$^{2}$,
Toby Jack$^{2}$,
Max Yang$^{2}$,
Haoran Li$^{3}$,
Nathan F. Lepora$^{2}$%
}

\begin{document}

\twocolumn[{%
\begin{@twocolumnfalse}

\maketitle
\vspace{-10mm}

\begin{center}

\setlength{\tabcolsep}{2pt}
\begin{tabular}{@{}c@{\hspace{-0.1mm}}c@{}}
    \includegraphics[width=0.565\textwidth]{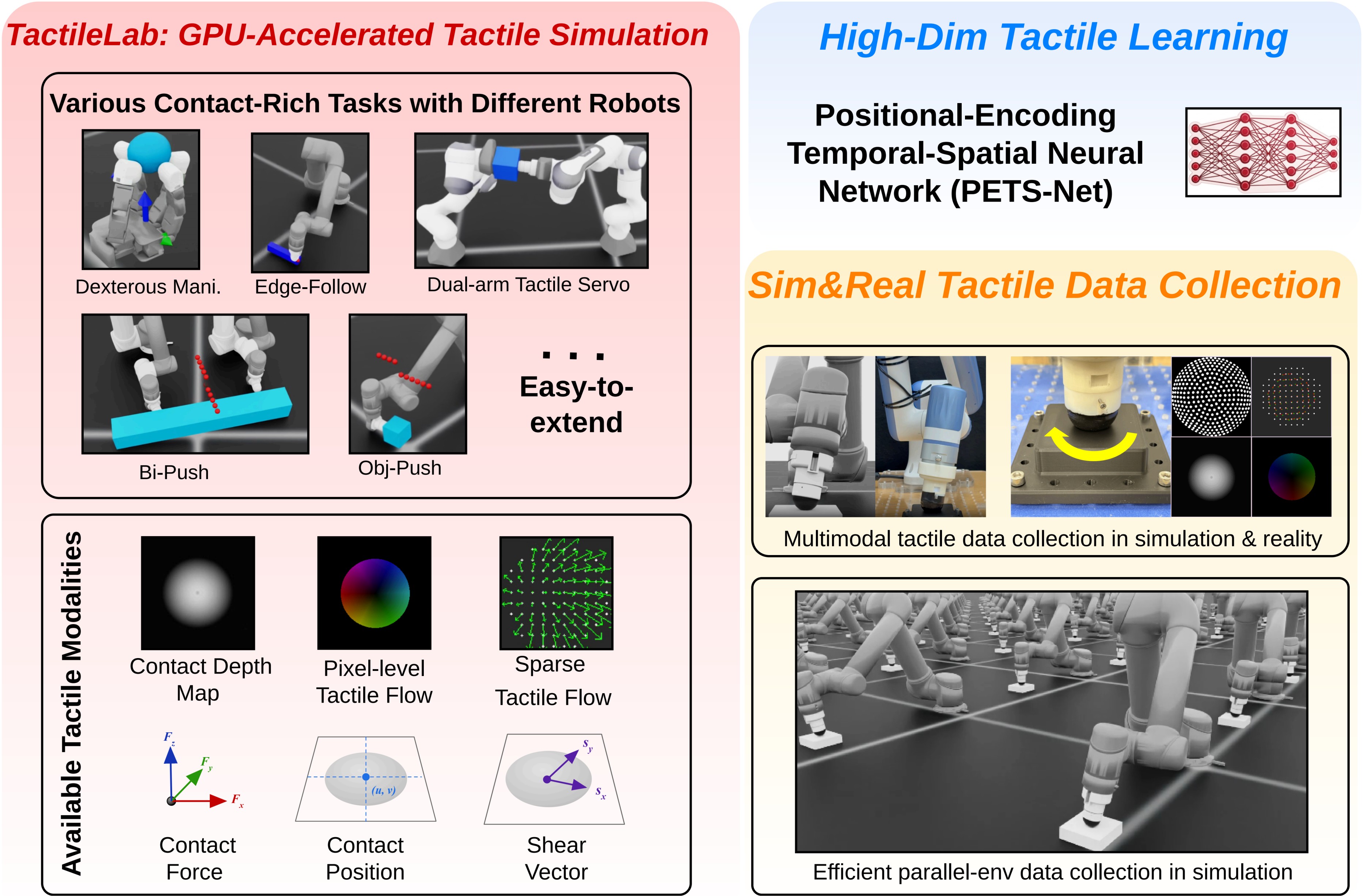} &
    \includegraphics[width=0.435\textwidth]{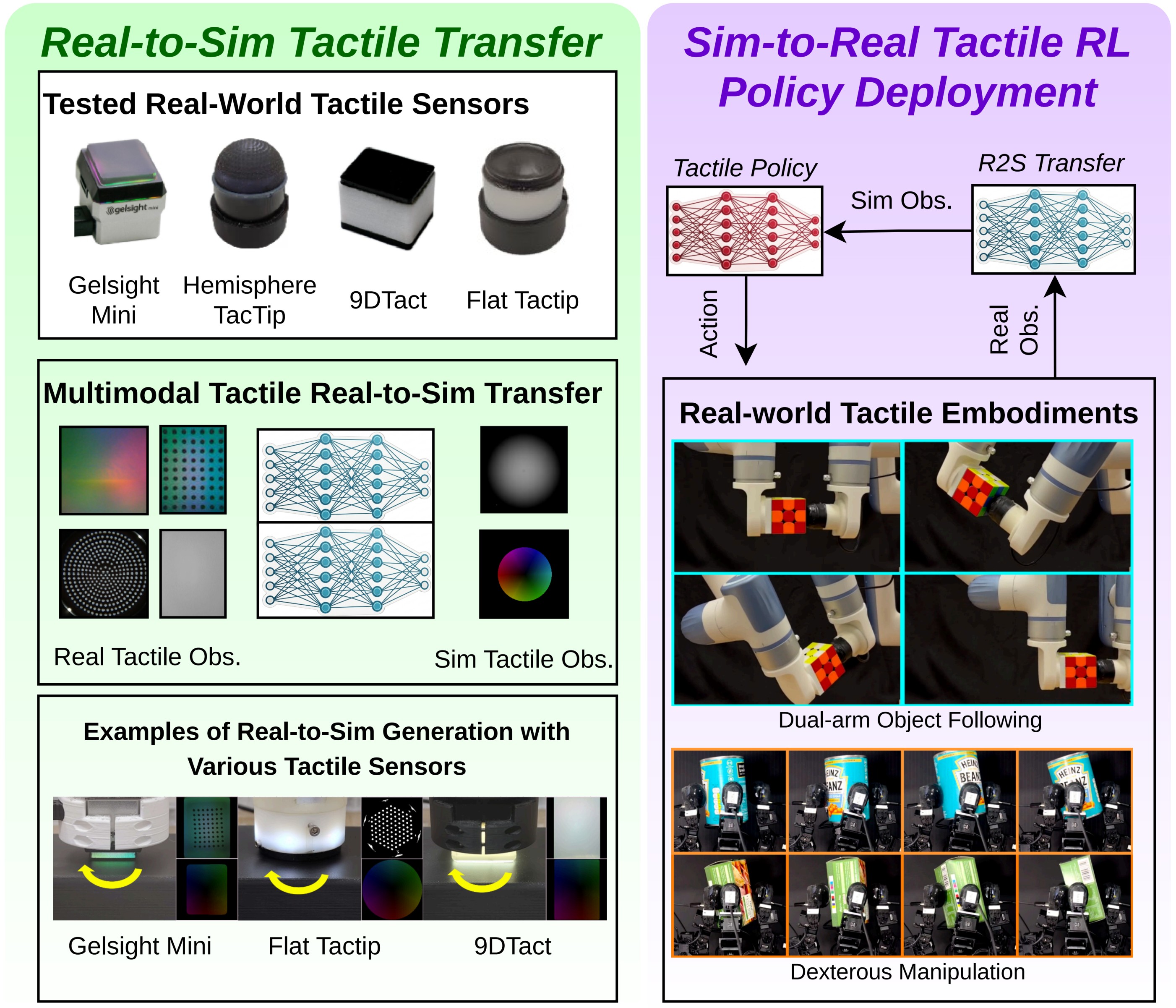}
\end{tabular}

\vspace{-0.5mm}

\captionof{figure}{\small TactileLab is an end-to-end framework for efficient, shear-sensitive tactile robot learning. Built on IsaacLab, it integrates GPU-accelerated tactile simulation with a novel sensing modality, \textit{simulated tactile flow}, for diverse contact-rich tasks and heterogeneous robotic systems. The framework includes modules for tactile policy and representation learning, a real-to-sim transfer pipeline supporting several real tactile sensors, and sim-to-real deployment of tactile reinforcement learning policies for various robots. This unified design enables scalable policy training, efficient collection of simulated tactile data, and effective transfer from simulation to real-world contact-rich manipulation.}
\label{fig:overview}

\end{center}
\end{@twocolumnfalse}
}]

\begingroup
\renewcommand{\thefootnote}{}
\footnotetext[0]{%
YL and NFL were supported by ARIA Robot Dexterity award SMRB-PR01-P19.
YL conducted this work at the University of Bristol.

$^{1}$OpenDIM Lab, School of Aerospace Engineering, Beijing Institute of Technology, Beijing, China.

$^{2}$Department of Engineering Mathematics and Bristol Robotics Laboratory, University of Bristol, Bristol BS8 1UB, U.K.

$^{3}$Xi'an Jiaotong-Liverpool University, Suzhou, China.
}
\endgroup
\begin{abstract}

Robotic simulators that represent shear within tactile sensing are becoming increasingly important for contact-rich manipulation. However, simulating contact shear with high-fidelity deformable or soft-body simulation is computationally demanding, limiting scalability to large-scale reinforcement learning. Here we introduce \textit{TactileLab}, a scalable multimodal tactile simulation framework built on IsaacLab, with a particular focus on a new optical-flow-based tactile representation termed \textit{simulated tactile flow}. Instead of relying on soft-body physics to model shear deformation, TactileLab uses rigid-body simulation to efficiently estimate dense tangential motion at the contact interface, producing tactile flow fields that encode shear-relevant contact dynamics. This flow field complements geometric features from a contact depth field. We further introduce a novel policy architecture for jointly processing long-horizon proprioceptive signals and rich tactile observations. Across multiple contact-rich tasks, policies trained using simulated tactile flow learn efficiently at scale and exhibit emergent contact-aware behaviours. To facilitate sim-to-real policy deployment, we then propose a novel real-to-sim tactile translation method that maps real tactile image streams to simulated tactile flow and depth fields, together with a new tactile data collection method, and validate it across diverse tactile sensor designs. Using this pipeline, we demonstrate real-world deployment on tasks encompassing dual-arm and dexterous in-hand manipulation. Overall, TactileLab provides an end-to-end sim-to-real framework for scalable tactile policy learning, enabling diverse shear-sensitive manipulation across heterogeneous tactile robotic systems.\\ Project webpage: \url{https://yijionglin.github.io/tactilelab/}

\end{abstract}


\section{INTRODUCTION}\label{sec:Intro}


Tactile sensing provides local, contact-grounded feedback that is essential for fine manipulation tasks. This is the case when a robot must finely regulate interaction forces such as to maintain contact or respond to small changes at the contact interface. For such tasks, vision-based tactile sensors are attractive because they provide high-resolution and highly-sensitive spatial observations. However, scalable simulation methods still struggle to represent tangential interaction efficiently, which is needed for tasks that depend on shear forces.

Most existing tactile simulators fall into two categories: 1)~{\em soft-body simulators} can model compliant contact with high fidelity~\cite{si2024difftactile, zhang2026simtac, li2026taccel}, but can be too slow for large-scale reinforcement learning; 2) {\em rigid-body simulators} are fast and scalable, but usually provide only depth-like signals~\cite{church2022tactile,lin2022tactilegym2,gomes2021generation, wang2022tacto, luu2023simulation}. As a result, they are well suited for normal-contact reasoning but less effective for tasks in which shear, slip, or tangential motion matters. This limitation is important for manipulation problems such as in-contact trajectory following, insertion, and in-hand object manipulation in the presence of external perturbations. Depth-only observations cannot fully capture contact dynamics over time: similar depth signals may correspond to different contact states and future outcomes, making the task partially observable and harder for policies to learn. Because of this, recent tactile simulators have modelled distributed normal and shear feedback by relying on geometry-dependent contact formulations and object-specific preprocessing~\cite{akinola2025tacsl,dang2026hydroshear}. Here we introduce a new way to instead calculate dense shear cues directly from rigid-body simulation without those assumptions, using a novel optical-flow-based tactile representation that we term \textit{simulated tactile flow}. 

This new flow representation allows us to address the gap of efficient shear-sensitive tactile simulation and robot learning, by introducing \emph{TactileLab}: a complete sim-to-real framework that augments rigid-body tactile sensing with a dense tangential displacement field. Rather than explicitly simulating elastomer deformation, the framework represents contact using two complementary components: normal-contact depth and a tangential-displacement vector field. This combination gives a compact and sim-to-real transferable tactile observation that is efficient to compute, maps onto various tactile sensor designs, and is suitable for large-scale reinforcement learning.

Our main contributions are as follows:


\noindent 1. \textbf{A scalable multimodal tactile simulation framework for contact-rich manipulation}: 
TactileLab is built on top of IsaacLab~\cite{mittal2025isaaclab} and supports GPU-parallel simulation for efficient reinforcement learning. It provides multiple tactile sensing modalities, including contact forces, contact positions, contact depth images, and simulated tactile flow, allowing researchers to study both low-dimensional and high-dimensional tactile feedback within a unified framework.

\noindent 2. \textbf{A novel simulated tactile modality for shear-relevant contact interaction}: 
We introduce simulated tactile flow, an optical-flow-inspired tactile modality that represents dense tangential motion at the contact interface. Together with contact depth, simulated tactile flow provides complementary high-dimensional tactile observations of normal contact geometry and tangential contact motion, enabling policies to exploit shear-relevant cues without explicit soft-body simulation.

\noindent 3. \textbf{A diverse suite of tactile robotic manipulation environments}: 
TactileLab provides a set of contact-rich robotic environments for single-arm, dual-arm, and dexterous in-hand manipulation. These show the scalability and generality of TactileLab for policy learning with high-dimensional contact-depth and tactile flow observations. We conduct detailed simulation studies on two representative tasks: dual-arm object following and dexterous in-hand object manipulation.

\noindent 4. \textbf{A new sample-efficient policy architecture for fusing proprioceptive history and tactile observations}: 
We propose PETS-Net (Positional-Encoding Temporal-Spatial Network) for encoding long-horizon low-dimensional proprioceptive histories and fusing them with high-dimensional tactile observations. 
Experimental results show that PETS-Net learns more efficiently than baseline policy architectures, demonstrating the benefit of jointly encoding temporal robot-state histories and spatial tactile features for contact-rich manipulation.

\noindent 5. \textbf{A new real-to-sim tactile transfer method for multimodal tactile observations}: 
Building on a previous static transfer method for single tactile images, we propose separate real-to-sim translators for contact depth and tactile flow. Specifically, the tactile-flow translator takes real tactile image streams as input and predicts simulated tactile flow, extending real-to-sim transfer from static contact-depth reconstruction to dynamic tangential-motion representations. We further validate the real-to-sim pipeline across diverse tactile sensor designs, demonstrating its applicability beyond a single sensor configuration.

\noindent 6. \textbf{Real-world deployment of TactileLab-trained policies}: 
Using the proposed multimodal tactile simulation and real-to-sim transfer pipeline, we demonstrate successful real-world deployment of policies trained in TactileLab on two representative contact-rich manipulation tasks: dual-arm object following and dexterous in-hand object manipulation. These results demonstrate the applicability of the TactileLab framework across distinct robotic embodiments and manipulation settings.

TactileLab will be released as an open-source framework to lower the barrier to entry to tactile robot learning, enable reproducible research, and accelerate progress in tactile robotics.

\section{RELATED WORK} \label{sec:related}

\subsection{Tactile Simulation}
\label{subsec:simulation}

Existing tactile simulators broadly separate into rigid-body and deformable-body approaches. Rigid-body methods, including Tactile Gym~\cite{church2022tactile,lin2022tactilegym2}, TACTO~\cite{wang2022tacto}, and other large-scale frameworks~\cite{gomes2021generation,luu2023simulation}, generate tactile signals from contact geometry, penetration depth, or contact forces available in conventional physics engines. They support efficient contact simulation but generally represent instantaneous normal contact without explicitly modelling shear deformation. In contrast, soft-body simulators model marker or elastomer deformation using analytical or calibrated approximations, FEM, or material-point methods~\cite{si2022taxim, chen2023tacchi, si2024difftactile, du2024tacipc, zhao2024fots, li2026taccel, xu2026etac, chen2026univtac, zhang2026simtac}. Although they capture richer contact information, they require sensor-specific deformation models and calibrated material parameters. Full soft-body solvers are also computationally intensive and generally offer lower throughput than rigid-body simulation~\cite{chen2026univtac, zhang2026simtac}, limiting large-scale parallel data generation for data-hungry methods such as RL. Recent geometry-aware approaches additionally model shear-sensitive signals. TacSL~\cite{akinola2025tacsl} computes dense normal and shear force fields, while HydroShear~\cite{dang2026hydroshear} models path-dependent hydroelastic shear. However, both require prior access to the contacting object shape through a signed-distance-function (SDF) representation and produce modelled force or marker-displacement fields. TactileLab instead derives tactile flow directly from rigid-body contact kinematics without object-specific SDF preprocessing, retaining scalable simulation while providing dense tangential-motion observations for shear-sensitive tasks.

A central design goal of TactileLab is to support both traditional low-dimensional tactile signals and rich high-dimensional tactile observations. Low-dimensional signals, such as contact positions, contact forces, and shear vectors, provide compact information about contact events and interaction dynamics~\cite{xu2022efficient, yang2023sim2real, chen2023tacchi, yang2024anyrotate, si2024difftactile, freud2025simshear}. In contrast, high-dimensional tactile observations, such as contact-depth images~\cite{gomes2021generation, church2022tactile, wang2022tacto, lin2022tactilegym2, akinola2025tacsl} and optical-flow-based tactile flow~\cite{du2021otf2, zhang2022otf3densecolor, funk2024evetac}, provide dense spatial information over the tactile sensing surface. This allows researchers to study how different levels of tactile feedback affect policy learning, from compact contact-state representations to image-like tactile observations suitable for end-to-end learning.

\subsection{Tactile Sim-to-Real Policy for Robotic Manipulation}
Early tactile sim-to-real policies were mainly developed for robotic arms equipped with vision-based tactile end-effectors. For example, the Tactile Gym series trains RL policies using simulated contact-depth images and maps real tactile images into the same observation space through paired real-to-sim translation~\cite{church2022tactile,lin2022tactilegym2, yang2023sim2real, comi2024touchsdf, wu2025vitacgen, lin2026neuraltouch}. These methods bridge the tactile observation gap effectively, but represent instantaneous contact geometry rather than tangential contact motion. 

With the development of more advanced tactile simulators, recent methods~\cite{si2024difftactile,akinola2025tacsl} have further demonstrated sim-to-real transfer for contact-rich manipulation using richer or more physically grounded tactile observations. More closely related to tangential contact sensing, approaches~\cite{zhao2024fots,dang2026hydroshear,li2026taccel} incorporate shear-related tactile observations into policy learning and demonstrate sim-to-real transfer on manipulation tasks. However, such tactile sim-to-real approaches have so far been primarily demonstrated on robotic arms or grippers, while dexterous-hand manipulation has generally relied on proprioception or lower-dimensional contact representations.

At the dexterous-hand level, rapid motor adaptation first enabled zero-shot in-hand rotation by inferring latent object properties from proprioceptive history~\cite{qi2023inhand}. RotateIt extends this proprioception-only framework by incorporating visual observations and discretized tactile information~\cite{qi2024inhand}. Touch Dexterity instead transfers in-hand rotation policies using binary contact taxels distributed over the fingers and palm~\cite{yin2023rotating}. More recent methods use compact contact representations: AnyRotate estimates contact pose and force from real tactile images during tactile sim-to-real policy deployment~\cite{yang2024anyrotate}, while Pan et al.~\cite{pan2026beyond} introduce a physics-grounded centre-of-pressure representation that captures contact force and location for in-hand manipulation.

Collectively, these studies demonstrate effective tactile sim-to-real transfer using depth images, binary or discretized contact signals, compact contact properties, and shear-related representations. However, a common dense temporal tactile representation that can support sim-to-real transfer across different types of vision-based tactile sensors and robotic embodiments remains underexplored. TactileLab introduces a real-to-sim transfer method that maps consecutive real tactile images into the dense tactile-flow space used during simulation, supporting policy transfer across different sensor designs, robotic embodiments, and manipulation tasks.

\begin{figure*}[t]
    \centering
    \includegraphics[width=\textwidth]{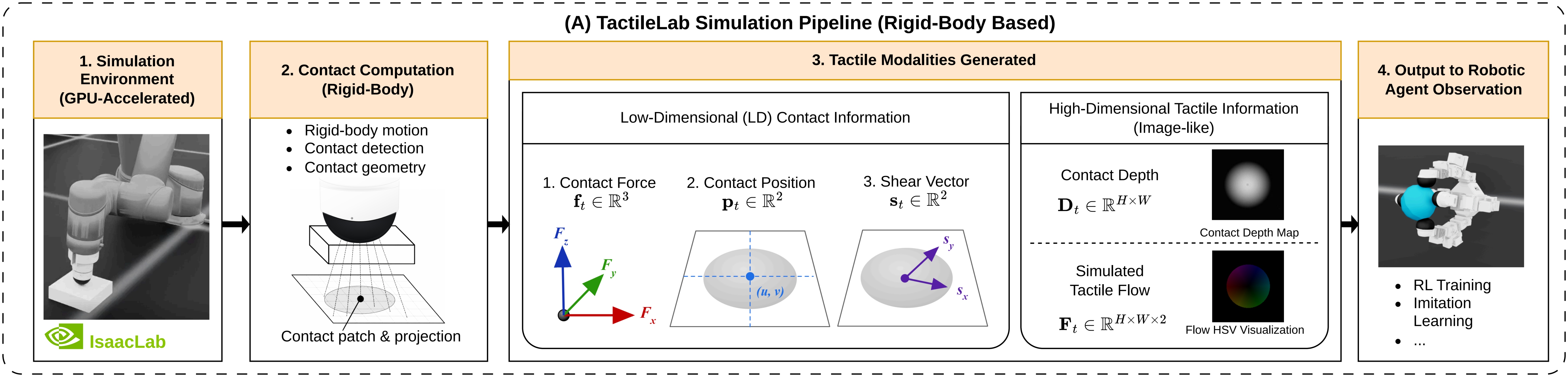}
    \par\nointerlineskip
    \includegraphics[width=\textwidth]{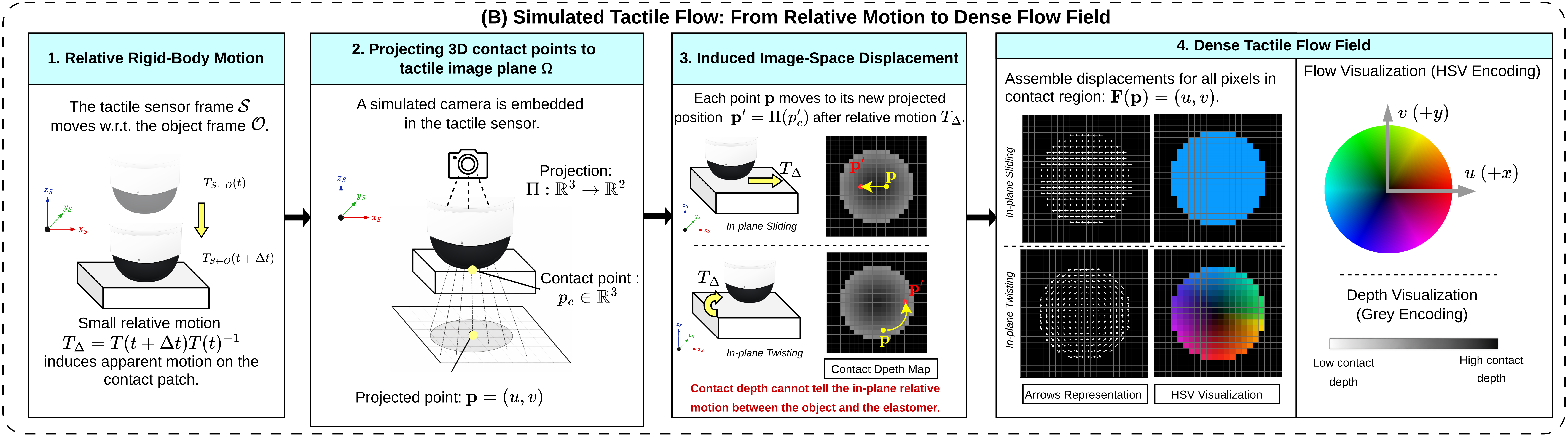}
    \caption{\small Overview of the TactileLab simulation framework and simulated tactile-flow computation. The rigid-body simulation pipeline computes contact geometry and generates low-dimensional contact descriptors, contact-depth images, and dense tactile-flow fields for policy learning. Simulated tactile flow is obtained by projecting the three-dimensional contact patch onto the tactile image plane and calculating the image-space displacement induced by the relative motion between the tactile sensor and the contacted object.}
    \label{fig:tactile_illustration}
    \vspace{-1em}
\end{figure*}
\section{Multimodal Tactile Robotic Simulator} \label{sec:method}
\subsection{Overview of TactileLab} \label{subsec:tactilelab_overview}

TactileLab is a scalable multimodal tactile simulation framework built on top of IsaacLab~\cite{mittal2025isaaclab}. Rather than pursuing high-fidelity soft-body simulation of tactile elastomers, TactileLab focuses on the efficient generation of tactile observations within rigid-body simulation, enabling large-scale reinforcement learning for contact-rich robotic manipulation.  In particular, TactileLab introduces simulated tactile flow, which provides a dense representation of tangential contact motion, complementing contact depth without requiring computationally expensive soft-body simulation. While instantiated here using rigid-body contact kinematics, tactile flow itself is not inherently tied to rigid-body physics and can naturally be extended to deformable-body simulation by tracking the motion of points on the tactile elastomer surface. 


TactileLab follows the design desiderata of efficient vision-based tactile simulation, including high throughput, flexibility, and ease of use.  Similar to prior efficient tactile simulation approaches~\cite{church2022tactile}, TactileLab does not aim to reproduce the full physical deformation or optical transduction process of a specific tactile sensor. Instead, it simulates task-relevant tactile quantities that can be generated efficiently at scale and later aligned with real tactile observations through real-to-sim transfer. It prioritizes scalable data generation, compatibility with large-scale reinforcement learning, and flexibility across sensor morphologies and manipulation tasks.



\subsection{Multimodal Tactile Simulation} \label{subsec:Multimodal_Tactile}

TactileLab provides multiple tactile sensing modalities within the same simulation framework. At each simulation step, a rigid-body simulator can provide the robot state, object state, contact information, and relative motion between the tactile sensor and contacting objects. Based on this information, TactileLab generates observations at different levels of abstraction over low- and high-dimensional tactile information. 

Low-dimensional tactile information $\mathbf{c}_t$ includes compact contact descriptors such as contact force, position, and shear over time $t$ (Fig.~\ref{fig:tactile_illustration} A3, left). These signals summarize contact interaction through physically meaningful quantities that are useful for analysis and privileged learning. In contrast, high-dimensional tactile observations such as contact depth and tactile flow are represented as dense fields over the tactile sensing surface, which are closer to the image-like observations produced by vision-based tactile sensors.

High-dimensional tactile information includes the contact-depth image $\mathbf{D}_t \in \mathbb{R}^{H \times W}$ at time~$t$, where each pixel value in the $(H,W)$-dimensional array represents the local normal contact depth at a corresponding point on the tactile sensing surface (Fig.~\ref{fig:tactile_illustration} A3, right). This provides a dense representation of normal contact geometry, indicating where contact occurs and how deeply the object indents into the sensor surface. Likewise, the tactile-flow field is denoted by $\mathbf{F}_t \in \mathbb{R}^{H \times W \times 2}$, where each pixel has a two-dimensional tangential motion vector at the contact interface. 
Together, these constitute a multimodal tactile observation
\[
    \mathbf{o}^{\mathrm{tac}}_t =
    \left\{
    \mathbf{D}_t,\,
    \mathbf{F}_t,\,
    \mathbf{c}_t
    \right\},
\]
where $\mathbf{D}_t$ denotes the contact-depth image, $\mathbf{F}_t$ the tactile-flow field, and $\mathbf{c}_t$ denotes low-dimensional contact information, including contact force, position, and shear vectors. 

In practice, different tasks and experiments may use different subsets of these modalities. For example, contact-depth images are used across the broader task suite to demonstrate scalable tactile policy learning, while contact depth and tactile flow are used in representative shear-relevant tasks to study multimodal tactile feedback and real-world deployment.

\subsection{Simulated Tactile Flow}\label{subsec:simulated_tactile_flow}


Vision-based tactile sensors capture both spatial contact geometry and temporal cues from in-contact motion, such as slip, stick--slip transitions, and tangential displacement. In real sensors, these cues are observed through optical flow or marker motion in tactile image sequences~\cite{du2021otf2,zhang2022otf3densecolor,funk2024evetac}. Simulating them remains challenging: soft-body methods can reproduce elastomer deformation and realistic image motion but are computationally expensive, whereas rigid-body simulators scale efficiently to parallel reinforcement learning but generally lack a dense representation of tangential contact motion.

To bridge this gap, we introduce \emph{simulated tactile flow}: an optical-flow-inspired tactile modality that represents dense in-plane motion at the contact interface. To the best of our knowledge, this is the first work to formulate dense contact motion as a high-dimensional tactile observation for policy learning in scalable rigid-body simulation. The simulated tactile flow is constructed directly from contact kinematics while providing an efficient, object-agnostic representation of shear-relevant contact cues. The details are described below.

\textit{a) Tactile Flow Representation:} At time $t$, the simulated tactile flow is defined as a 2D vector field over a tactile image
\[
\mathbf{F}_{t}(\mathbf{p})
=
\begin{bmatrix}
F_{t}^{u}(\mathbf{p})\\
F_{t}^{v}(\mathbf{p})
\end{bmatrix}\in
\mathbb{R}^{H\times W\times 2},\ \ 
\mathbf{p}=(u,v)\in\Omega\subset\mathbb{Z}^2,
\]
where $u \in \{0,\ldots,W-1\}$ and $v \in \{0,\ldots,H-1\}$. Here $F_{t}^{u}(\mathbf{p})$ and $F_{t}^{v}(\mathbf{p})$ denote the
horizontal and vertical image-space displacements, respectively,
between times $t$ and $t+\Delta t$.
Each tactile pixel is associated
with a two-dimensional motion vector. The resulting dense tactile-flow field is compatible with convolutional policy networks and can be spatially aligned with contact-depth images for multimodal fusion.

Although simulated tactile flow is represented in image coordinates, it is induced by relative tangential motion between the contacting object and the tactile sensor. It thus provides an optical-flow-inspired representation of shear-relevant contact cues without explicitly simulating elastomer deformation. Importantly, the proposed tactile-flow representation is independent of the underlying contact dynamics formulation. Although TactileLab computes it from rigid-body contact kinematics for efficiency and scalability, the same representation can also be derived from deformable-body simulations by tracking the motion of surface points at the sensor interface.

\textit{b) Rigid-Body Computation for Tactile Flow:}
In TactileLab, simulated tactile flow is computed as the dense
image-space displacement induced by relative rigid-body motion between
the contacting object and the tactile sensor. The tactile sensor is modeled as an embedded camera observing the contacting object surface through the tactile skin, where the resulting contact geometry is approximated by the geometric penetration of the object surface into the nominal tactile-elastomer surface. Given a contact-depth map $\mathbf{D}_{t}(\mathbf{p})$ over pixels $\mathbf{p}\in\Omega$, we define a binary contact mask
\[
\mathbf{C}_t(\mathbf{p})
=
\begin{cases}
1, & \mathbf{D}_{t}(\mathbf{p})>\epsilon,\\
0, & \text{otherwise},
\end{cases}
\]
where $\epsilon$ is a small contact-depth threshold. For each contact pixel $\mathbf{p}$ satisfying $C_t(\mathbf{p}) = 1$, let $\mathbf{x}_t(\mathbf{p}) \in \mathbb{R}^3$ denote the corresponding point on the contacting object surface that penetrates the nominal tactile-elastomer surface, expressed in the tactile-sensor frame. Its image coordinate is then given by $\mathbf{p} = \Pi(\mathbf{x}_t(\mathbf{p}))$, where $\Pi : \mathbb{R}^3 \rightarrow \mathbb{R}^2$ denotes the ray projection onto the tactile image plane.

The simulated tactile-flow field over the complete tactile-image domain
is then defined as
\[
\mathbf{F}_{t}(\mathbf{p})
=
\begin{cases}
\widetilde{\mathbf{F}}_{t}(\mathbf{p}),
& \mathbf{C}_t(\mathbf{p})=1,\\[3pt]
\mathbf{0},
& \mathbf{C}_t(\mathbf{p})=0,
\end{cases}
\]
over points $\bm p \in \Omega$. Here the raw image-space displacement is
\begingroup
\small
\[
\widetilde{\mathbf{F}}_{t}(\mathbf{p})
=
\Pi\!\left(\mathbf{x}_{t+\Delta t}(\mathbf{p})\right)
-
\Pi\!\left(\mathbf{x}_{t}(\mathbf{p})\right),\ \ \ 
\mathbf{p}=\Pi(\mathbf{x}_{t}(\mathbf{p})),
\]
\endgroup
where $\mathbf{x}_{t+\Delta t}(\mathbf{p})$ denotes the same object-surface point after being transformed by the relative rigid-body motion over one simulation timestep, expressed in the tactile-sensor frame. The resulting $\mathbf{F}_{t}\in\mathbb{R}^{H\times W\times 2}$ is the dense tactile-flow representation defined in part~a), with zero flow outside the contact region. It is computed directly from rigid-body motion, contact geometry, and camera projection, without explicitly simulating tactile-elastomer deformation.

\textit{c) Relation to Contact Depth:}
Contact-depth images provide a clean and physically interpretable representation of normal contact geometry. They indicate where contact occurs and how deeply the object presses into the tactile surface, making them effective for many tactile manipulation tasks. However, contact depth primarily captures normal interaction and does not explicitly encode tangential contact motion.

This limitation becomes critical for tasks where shear, slip, or persistent in-contact motion play a central role. Two contact states may produce similar depth images for distinct tangential motion patterns. Simulated tactile flow complements contact depth by providing dense in-plane motion cues at the contact interface. While contact depth represents where contact occurs, tactile flow represents how the contact is moving tangentially.

Together, contact depth and tactile flow provide complementary high-dimensional tactile observations of normal contact geometry and tangential contact motion. This enables TactileLab to represent shear-relevant interaction cues without explicitly simulating soft-body deformation, while remaining efficient enough for large-scale reinforcement learning. 



\section{Multimodal Tactile Policy Learning}
\label{sec:multimodal_tactile_policy_learning}

\begin{figure*}
  \centering
     \includegraphics[width=1\linewidth]{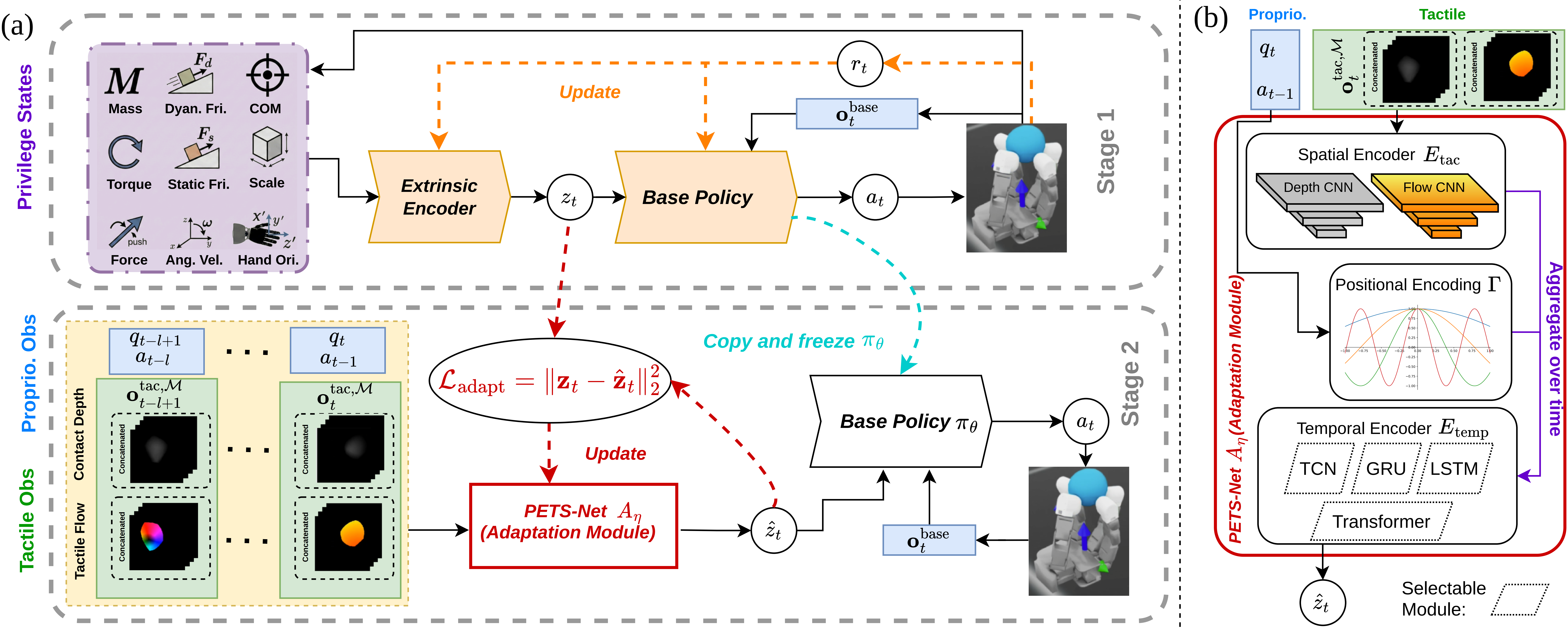}

    \caption{\small (a) Overview of multimodal tactile policy learning for dexterous in-hand manipulation. To evaluate the benefit of our rich tactile observations, unknown random force and torque perturbations are applied to the manipulated object, making the task more challenging and requiring robust and informative tactile feedback. (b) The structure of the proposed PETS-Net for high-dimensional tactile policy learning.}
  \label{fig:multimodal_policy_learning}
\end{figure*}

\subsection{Task Setting and Multimodal Tactile Observation}

We study multimodal tactile policy learning on two representative contact-rich tasks: dual-arm object-following and dexterous in-hand object manipulation (as shown in Fig.~\ref{fig:overview}). Both tasks require the policy to reason about contact evolution over time and to react to normal and tangential interaction at the tactile interface. The two tasks share a similar teacher--student learning structure (Fig.~\ref{fig:multimodal_policy_learning}). The task descriptions are as follows while more details are given in Sec.~\ref{subsec:task_describe_dual_arm} and ~\ref{subsec:task_describe_in_hand}: 

\indent 1) Dual-arm object-following in a 6D workspace is a shear-dependent task that has previously been studied using tactile sensing and a traditional PID controller tuned in the real world~\cite{lloyd2024shear}. The goal is for a tactile follower arm to maintain stable contact with an object moved by a leader arm along a predefined trajectory. Here, we investigate whether simulated tactile flow enables this task to instead be learned with deep RL in the TactileLab simulation. The action space consists of 6D pose changes of the tactile follower arm.\\
\indent 2) The dexterous in-hand object manipulation task follows settings similar to those of prior works~\cite{qi2023inhand,yang2024anyrotate}, in which a policy controls a dexterous hand to rotate an object held by the fingers without palm support. The action space consists of incremental commands for the 16 finger-joint angles. To further assess the benefits of richer tactile modalities, the manipulation environment is made more challenging by applying unknown random force and torque perturbations to the object.

The simulator dynamics define an underlying finite-horizon Markov decision process (MDP) with state \(\mathbf{s}_t\), action \(\mathbf{a}_t\), reward \(r_t\), and discount factor \(\gamma\). However, the deployable policy receives only partial observations \(\mathbf{o}_t\), which appear as a partially observable Markov decision process (POMDP). Consequently, the base policy uses a short proprio-action history to provide local velocity- and acceleration-related information through recent proprioceptive states and action changes
\[
\mathbf{o}^{\mathrm{base}}_t =
\left[
\mathbf{q}_{t-L+1:t},
\mathbf{a}_{t-L:t-1}
\right],
\]
where \(\mathbf{q}_t\) denotes the proprioceptive states (such as the hand joint configuration or the robot arm end-of-effector's pose), \(\mathbf{a}_{t-1}\) is the previous action, and \(L\) is the short history length used by the base policy.

In the experiments, the student observation is selectable across different modality combinations. 
The set of available tactile modalities includes some image-like modalities
\setlength{\abovedisplayskip}{3pt}
\setlength{\belowdisplayskip}{3pt}
\[
\mathcal{T}
=
\{\mathrm{depth},\mathrm{flow},\mathrm{contact}\}
\supset
\mathcal{T}^{\mathrm{img}}
=
\{\mathrm{depth},\mathrm{flow}\}.
\]

\noindent Then $\mathcal{M}
\subseteq
\mathcal{T}$
denotes the subset of tactile modalities used in a particular experiment, with  $\mathcal{M}^\mathrm{img}=\mathcal{M}\cap\mathcal{T}^\mathrm{img}$ being image-like. 
Following the notation from Sec.~\ref{subsec:Multimodal_Tactile}, the corresponding tactile observation is written as
\(
\mathbf{o}^{\mathrm{tac},\mathcal{M}}_t
=
\left\{
\mathbf{o}^m_t
\mid
m \in \mathcal{M}
\right\},
\)
where
$\mathbf{o}^{\mathrm{depth}}_t = \mathbf{D}_t$,
 $\mathbf{o}^{\mathrm{flow}}_t = \mathbf{F}_t$, 
$\mathbf{o}^{\mathrm{contact}}_t = \mathbf{c}_t$ denote the observation components.
The deployable student system therefore uses the short base-policy input \(\mathbf{o}^{\mathrm{base}}_t\) together with a longer history \(\mathbf{o}^{\mathrm{tac},\mathcal{M}}_{t-l+1:t}\) for latent adaptation, where \(l\) is the adaptation history length.


\subsection{Two-stage Multimodal Tactile Policy Learning}

Here we adopt a two-stage teacher-student learning framework~\cite{qi2023inhand} (see Fig.~\ref{fig:multimodal_policy_learning}). In Stage 1, a teacher policy is trained in simulation using information unavailable on the real robot: $\mathbf{e}_t$ denotes the privileged extrinsic state, including the object's mass, pose, twist, friction, scale; the external perturbation forces and torques applied to the object; and the hand orientation.
The base policy  \(\pi_{\theta}\) then predicts actions with the short proprio-action history 
\(\mathbf{a}_t=\pi_{\theta}(\mathbf{o}^{\mathrm{base}}_t,\mathbf{z}_t)\).
using a privileged latent vector that is a compact representation mapped using an extrinsic encoder $
\mathbf{z}_t = E_{\psi}(\mathbf{e}_t)$. The teacher policy is optimized using RL to maximize the expected return \(J(\theta,\psi)=\mathbb{E}\!\left[\sum_{t=0}^{T}\gamma^t r_t\right]\).
This stage allows the policy to learn strong contact-rich behaviours with access to privileged simulation information. During student learning, the privileged extrinsic encoder is replaced by an adaptation module that estimates the extrinsic latent vector from real-world-available proprioceptive, action, and tactile-observation histories. 


In Stage 2, the base policy \(\pi_{\theta}\) is copied from Stage 1 and frozen, while the extrinsic encoder \(E_{\psi}\) estimates the latent vector \(\hat{\mathbf{z}}_t\) from histories of proprioceptive states, actions, and selected tactile observations available in the real world. The encoder is trained to match the privileged latent representation using a standard least-squares loss

\[
\mathcal{L}_{\mathrm{adapt}}
=
\left\|
\mathbf{z}_t - \hat{\mathbf{z}}_t
\right\|_2^2.
\]

The student policy then acts using the same base-policy input as the teacher policy but with extrinsic vectors predicted from the adaptation module,
\[
\mathbf{a}_t =
\pi_{\theta}
\left(
\mathbf{o}^{\mathrm{base}}_t,
\hat{\mathbf{z}}_t
\right),\ \ \ 
\mathbf{o}^{\mathrm{base}}_t =
\left[
\mathbf{q}_{t-L+1:t},
\mathbf{a}_{t-L:t-1}
\right].
\]
Here, \(L\) denotes the short proprio-action history length used by the frozen base policy. Because \(\hat{\mathbf{z}}_t\) is inferred only from real-world-available proprioceptive, action, and tactile histories, the resulting policy can be deployed on the real robot without access to privileged simulation states.

\subsection{Proposed Positional-Encoding Temporal-Spatial Network}

Neural networks are known to exhibit a bias toward learning low-frequency functions, which can limit their ability to represent rapidly varying input-output relationships~\cite{rahaman2019spectral,mildenhall2021nerf}. This issue is particularly relevant in contact-rich dexterous manipulation, where high-dimensional state--action spaces, discontinuous contact patterns, and contact-mode transitions can cause small changes in joint configuration, previous action, or contact state to require different corrections~\cite{rajeswaran2018learning,liang2024robust}.

Here we propose a novel \emph{Positional-Encoding Temporal-Spatial Network} (\emph{PETS-Net}) and use it as the adaptation module \(A_{\eta}\), as shown in Fig.~\ref{fig:multimodal_policy_learning}(b). Although a standard adaptation module~\cite{qi2023inhand} can also replace the privileged extrinsic encoder during student learning, its representational capacity could be limited when the input consists of long-horizon proprioceptive histories, action histories, and high-dimensional tactile-observation histories. PETS-Net is designed to improve this adaptation process by learning a richer tactile-proprioceptive representation from real-world-available observations. Specifically, PETS-Net estimates the privileged latent vector from real-world-available temporal information by combining positional-encoded joint-angle and action histories with spatially encoded tactile-observation histories. The \emph{positional-encoding} component is applied inside the adaptation module to robotic proprioceptive states (e.g., joint angles) and action sequences, enabling proprioceptive and action histories to be represented more expressively. The \emph{spatial} component encodes selected high-dimensional tactile observations, such as contact-depth images and tactile-flow fields, and the \emph{temporal} component aggregates the encoded tactile features together with the positional-encoded proprio-action histories over time.

For compactness, we denote the positional-encoded proprio-action feature used by PETS-Net as
\[
\mathbf{g}_t =
\left[
\Gamma(\mathbf{q}_t),
\Gamma(\mathbf{a}_{t-1})
\right],\ \ 
\Gamma(\bm b_t)
:=
\left[
\phi_0(\bm b_t),
\ldots,
\phi_{K-1}(\bm b_t)
\right],
\]
where $\bm b_t$ denotes $\bm q_t$ or $\bm a_{t-1}$.
Here, the positional encoding $\Gamma$
is defined by frequency-band encoding over $K$ components
\[
\phi_k(\bm b_t)
=
\left[
\sin(2^k\pi \bm b_t),
\cos(2^k\pi \bm b_t)
\right], 
\]
concatenating the encodings 
\(
\phi_k(\bm b_t)=
\left[\phi_k(b^1_t),\cdots, \phi_k(b^n_t)
\right]
\)
of the individual components $\mathbf{b}_t=(b^1_t,\cdots,b^n_t)$.

For high-dimensional tactile inputs, the selected image-like modalities are encoded by modality-specific spatial tactile encoders: $E_{\mathrm{depth}}$ for contact-depth map, and $E_{\mathrm{flow}}$ for tactile flow. Their latent embeddings are concatenated with the low-dimensional contact descriptors \(\mathbf{c}_t\), when used, to form the tactile feature
\[
\begin{gathered}
\mathbf{z}^{\mathrm{tac}}_t =
\left[
\mathbf{z}^{\mathrm{depth}}_t,\,
\mathbf{z}^{\mathrm{flow}}_t,\,
\mathbf{c}_t
\right], \\[2pt]
\mathbf{z}^{\mathrm{depth}}_t =
E_{\mathrm{depth}}\!\left(
\mathbf{o}^{\mathrm{depth}}_t
\right),
\qquad
\mathbf{z}^{\mathrm{flow}}_t =
E_{\mathrm{flow}}\!\left(
\mathbf{o}^{\mathrm{flow}}_t
\right).
\end{gathered}
\]
Here, $\mathbf{z}^{\mathrm{depth}}$ and $\mathbf{z}^{\mathrm{flow}}$ denote the latent vectors encoding contact-depth and tactile-flow information, respectively. A temporal encoder \(E_{\mathrm{temp}}\) then aggregates tactile features and positional-encoded proprio-action features
\[
\mathbf{z}^{\mathrm{temp}}_t =
E_{\mathrm{temp}}
\left(
\mathbf{z}^{\mathrm{tac}}_{t-\ell+1:t},
\mathbf{g}_{t-\ell+1:t}
\right),
\]
over an adaptation history window of length $\ell$.
Finally, PETS-Net $A_{\eta}$ predicts the latent interaction representation 
\[
\hat{\mathbf{z}}_t =
A_{\eta}
\left(
\mathbf{o}^{\mathrm{tac},\mathcal{M}}_{t-\ell+1:t},
\mathbf{q}_{t-\ell+1:t},
\mathbf{a}_{t-\ell:t-1}
\right)
=
E_{\mathrm{lat}}
\left(
\mathbf{z}^{\mathrm{temp}}_t
\right)
\]
through a latent head \(E_{\mathrm{lat}}\).
This design allows the adaptation module to reason over both the spatial structure of selected tactile images and their temporal evolution, while also using a richer positional-encoded representation of proprioceptive and action histories.

\section{Real-to-Sim Tactile Transfer}

\begin{figure*}[t]
  \centering
  \includegraphics[width=0.95\linewidth]{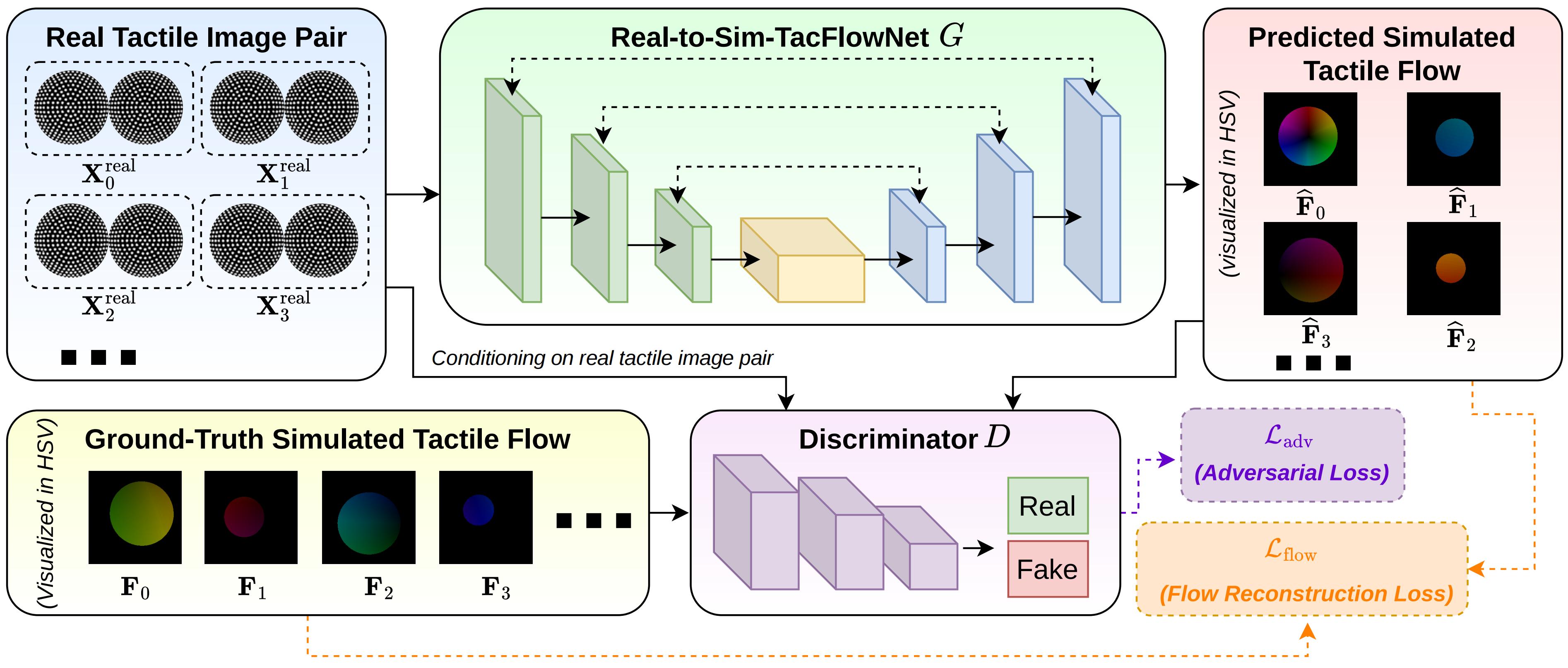}
  \vspace{-0.5em}
    \caption{\small Overview of Real-to-Sim-TacFlowNet. A pair of real tactile images is concatenated and passed to a pix2pix-style generator \(G\), which predicts a pixel-level simulated tactile-flow field. During training, the predicted flow is supervised by a paired simulated tactile-flow target through a reconstruction loss and by a discriminator \(D\) through an adversarial loss. The HSV flow images are used only for visualization; the network output is a two-channel flow field \(\widehat{\mathbf{F}}\).}
  \label{fig:tacflownet}
\end{figure*}

\subsection{Real-to-Sim Tactile-Flow Translation Formulation}\label{subsec:Real-to-Sim_tactile_flow}
To bridge the observation gap between real vision-based tactile sensing and the high-dimensional simulated tactile modalities used in our policy-learning framework, we formulate real-to-sim tactile-flow prediction as a pixel-level image-pair-to-image translation problem. Given two consecutive real tactile images, the model will predict a pixel-aligned simulated tactile-flow field 
that is directly compatible with the tactile modality available in TactileLab.

In this work, the high-dimensional real-to-sim tactile translation includes contact-depth maps and tactile-flow fields (see Sec.~\ref{subsec:simulated_tactile_flow}). Since contact-depth prediction from vision-based tactile images has been explored in prior work~\cite{church2022tactile, lin2022tactilegym2}, we use it as a supporting modality and focus on the more challenging problem of learning pixel-level simulated tactile flow from real tactile image pairs. Unlike single-frame contact-depth prediction~\cite{church2022tactile, lin2022tactilegym2}, which mainly estimates normal contact geometry from a single tactile image, our formulation predicts a dense temporal tactile-flow field from two consecutive real tactile images, capturing local in-plane motion caused by tangential interaction. It also differs from low-dimensional shear-vector estimation~\cite{lloyd2024shear,freud2025simshear}, as the output preserves pixel-level spatial structure across the tactile image.


Given two consecutive real tactile images
$\mathbf{I}_t,\mathbf{I}_{t+\Delta t}\in[0,1]^{H\times W}$,
we define the real-to-sim tactile-flow mapping
\(\Phi:[0,1]^{H\times W}\times[0,1]^{H\times W}
\rightarrow\mathbb{R}^{H\times W\times 2}\)
as a function that maps a pair of real tactile images to the
corresponding simulated tactile-flow field:
\begingroup
\setlength{\abovedisplayskip}{3pt}
\setlength{\belowdisplayskip}{3pt}
\[
\mathbf{F}_t := \Phi(\mathbf{I}_t,\mathbf{I}_{t+\Delta t}).
\]
\endgroup

\noindent Our aim is to learn a real-to-sim tactile transfer $G$ that approximates
$\Phi$, as described in the next subsection.
\subsection{TacFlowNet: Learning Pixel-Level Tactile-Flow Prediction}


Having defined the desired real-to-sim tactile-flow mapping $\Phi$ in
Sec.~\ref{subsec:Real-to-Sim_tactile_flow}, we now introduce Real-to-Sim-TacFlowNet (TacFlowNet), a learnable model $G$ (as shown in Fig.~\ref{fig:tacflownet}) that
approximates this mapping from real tactile-image pairs to simulated
tactile-flow fields. During training, each real-to-sim sample consists of a real start--final tactile image pair and its paired simulated tactile-flow target. Because the instantaneous real robot velocity during this short motion is unavailable and may be non-constant, we treat the start--final image pair as an effective one-step tactile transition corresponding to one tactile-sensor sampling interval. The detailed sensor-frequency normalization used to construct the paired simulated target is described in Sec.~\ref{subsec:Real-to-Sim_data_collection}.

We concatenate the two real tactile images along the channel dimension to form a single sample
\(\mathbf{X}^{\mathrm{real}}_i\)
so that the model can learn motion-dependent deformation cues from the tactile appearance change between the two contact states. The mapping to a simulated tactile-flow target \(\mathbf{F}_i\) is implemented by a convolutional generator \(G\), denoted as TacFlowNet: following a pix2pix-style conditional image-to-image translation formulation~\cite{isola2017image}, the network predicts

\begingroup
\small
\setlength{\abovedisplayskip}{4pt}  
\setlength{\belowdisplayskip}{4pt}
\[
\widehat{\mathbf{F}}_i
=
G(\mathbf{X}^{\mathrm{real}}_i)
=
G\!\left(
[\mathbf{I}^{\mathrm{real}}_{i,\mathrm{start}},
 \mathbf{I}^{\mathrm{real}}_{i,\mathrm{final}}]
\right),
\]
\endgroup

\noindent where the corresponding tactile flow target is defined in Sec.~\ref{subsec:simulated_tactile_flow}. 
The dense pixel-level supervision allows the model to learn a direct real-to-sim mapping from real tactile image pairs to the tactile-flow modality.


To encourage both pixel-level accuracy and realistic spatial flow structure, we train the mapping using a conditional adversarial objective together with a supervised flow reconstruction loss~\cite{isola2017image}. The discriminator \(D\) receives the real tactile image pair together with either the paired simulated tactile-flow target or the generated tactile-flow prediction. An adversarial loss

\begingroup
\small
\setlength{\abovedisplayskip}{4pt}
\setlength{\belowdisplayskip}{4pt}
\[
\begin{aligned}
\mathcal{L}_{\mathrm{adv}}(G,D)
={}&
\mathbb{E}_{\mathbf{X}^{\mathrm{real}}_i,\mathbf{F}_i}
\left[
\log D
\left(
\mathbf{X}^{\mathrm{real}}_i,
\mathbf{F}_i
\right)
\right]
\\[-2pt]
&+
\mathbb{E}_{\mathbf{X}^{\mathrm{real}}_i}
\left[
\log
\left(
1-
D
\left(
\mathbf{X}^{\mathrm{real}}_i,
\widehat{\mathbf{F}}_i
\right)
\right)
\right],
\end{aligned}
\]
\endgroup

\noindent is combined with a masked pixel-level flow reconstruction loss

\begingroup
\small
\setlength{\abovedisplayskip}{4pt}
\setlength{\belowdisplayskip}{4pt}
\[
\mathcal{L}_{\mathrm{flow}}(G)
=
\mathbb{E}_{\mathbf{X}^{\mathrm{real}}_i,\mathbf{F}_i}
\left[
\frac{
\left\|
\mathbf{C}_i \odot
\left(
\mathbf{F}_i-\widehat{\mathbf{F}}_i
\right)
\right\|_1
}{
\|\mathbf{C}_i\|_1+\delta
}
\right],
\]
\endgroup
where \(\odot\) denotes element-wise multiplication, \(\mathbf{C}_i\) denotes the binary contact mask associated with the \(i\)-th simulated tactile-flow target, and \(\delta>0\) prevents division by zero. This loss enforces dense agreement between the predicted and target simulated tactile-flow fields within the contact region. The overall training objective is parametrised by a coefficient \(\lambda_{\mathrm{flow}}\), 
\[
G^{\star},D^{\star}
=
\arg\min_G \max_D
\left(
\mathcal{L}_{\mathrm{adv}}(G,D)
+
\lambda_{\mathrm{flow}}
\mathcal{L}_{\mathrm{flow}}(G)
\right),
\]
that balances adversarial realism vs pixel-level flow accuracy.

The key contribution of TacFlowNet is the temporal, pixel-level prediction of simulated tactile flow from real tactile image pairs. 
Unlike analytical optical-flow pipelines that explicitly track printed markers, dense color patterns, or local image patches, our method learns a direct mapping from real tactile appearance changes to the simulated tactile-flow modality used for policy learning. The predicted output is therefore a simulated tactile-flow representation, rather than the true optical flow of the real tactile images. 

During real-world deployment, TacFlowNet receives consecutive tactile frames,
\(
\mathbf{X}^{\mathrm{real}}_t
=
\left[
\mathbf{I}^{\mathrm{real}}_{t-1},
\mathbf{I}^{\mathrm{real}}_t
\right],
\)
captured at the tactile-sensor sampling frequency. The trained generator \(G^{\star}\) converts each consecutive image pair into pixel-level simulated tactile-flow observations, enabling policies trained with simulated tactile flow to operate from real tactile inputs.


\section{Experiments}

\subsection{Real-to-Sim Tactile Data Collection}
\label{subsec:Real-to-Sim_data_collection}

To train the real-to-sim tactile translators, we collect paired real and simulated tactile observations using a shared contact-and-shear protocol. Each sample is specified by a 6D contact pose
\(
\boldsymbol{\xi}_i=(x_i,y_i,z_i,\alpha_i,\beta_i,\psi_i)\in\mathbb{R}^{6}
\)
and a prescribed relative motion \(\Delta\boldsymbol{\xi}_i\), where \((x_i,y_i,z_i)\) denotes the contact position and \((\alpha_i,\beta_i,\psi_i)\) denotes roll, pitch, and yaw. The sampling ranges for contact poses and relative motions are reported in the Appendix.

For each sample, the real tactile sensor is first moved to a clearance pose and then brought into contact with the object at \(\boldsymbol{\xi}_i\). After contact is established, we record the initial tactile image \(\mathbf{I}^{\mathrm{real}}_{i,\mathrm{start}}\). The sensor then executes the prescribed motion to \(\boldsymbol{\xi}_i+\Delta\boldsymbol{\xi}_i\), while maintaining contact, and the final tactile image \(\mathbf{I}^{\mathrm{real}}_{i,\mathrm{final}}\) is recorded. This yields a real tactile image pair
\[
\mathbf{X}^{\mathrm{real}}_i =
\left[
\mathbf{I}^{\mathrm{real}}_{i,\mathrm{start}},
\mathbf{I}^{\mathrm{real}}_{i,\mathrm{final}}
\right],
\]
capturing the change in elastomer deformation induced by the contact motion.

Corresponding simulation labels are generated using the tactile sensor model in TactileLab, based on the sampled contact poses and relative motions. We record contact-depth maps at the initial and final states, denoted by \(\mathbf{D}_{i,\mathrm{start}}\) and \(\mathbf{D}_{i,\mathrm{final}}\), together with the associated dense tactile-flow field \(\mathbf{F}_i\). The resulting paired dataset is
\[
\mathcal{D} =
\left\{
\left(
\mathbf{X}^{\mathrm{real}}_i,
\mathbf{D}_{i,\mathrm{start}},
\mathbf{D}_{i,\mathrm{final}},
\mathbf{F}_i
\right)
\right\}_{i=1}^{N}.
\]
These labels supervise the translation of real tactile observations into contact-depth and tactile-flow representations. During deployment, the trained translators process consecutive real tactile frames.

\begin{figure*}[t]
  \centering
  \includegraphics[width=1\linewidth]{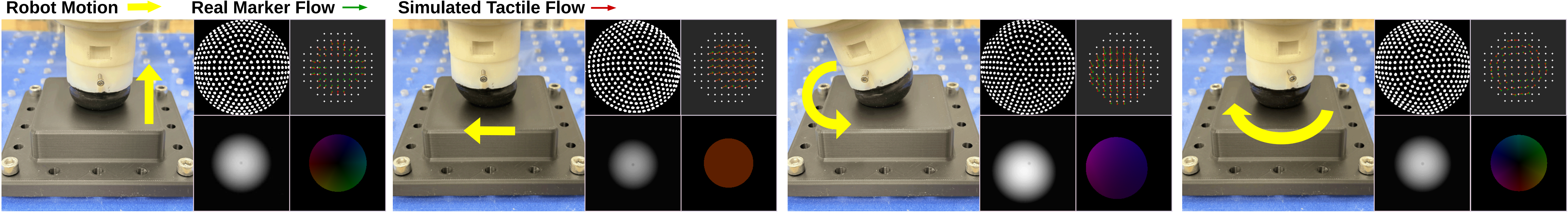}
  \vspace{-1em}
    \caption{\small Examples of paired real--sim tactile data collection for representative contact motions, including press, slide, tilt, and twist. Each example shows the real robot collection setup, corresponding final tactile image from the real image pair, real vs. generated sparse flow fields, simulated contact depth, and dense simulated tactile flow. 
    }
  \label{fig:example_Real-to-Sim_flow}
\end{figure*}

As shown in Fig.~\ref{fig:example_Real-to-Sim_flow}, the collected dataset covers representative contact motions, including pressing, sliding, tilting, and twisting. For visual reference, sparse marker flow is computed from the real tactile images using a classical optical-flow method, illustrating the observed shear deformation. The simulated tactile-flow fields provide dense pixel-level supervision for learning motion-related tactile features.

\subsection{Dual-Arm Object-Following Task Setup}
\label{subsec:task_describe_dual_arm}

To evaluate whether the proposed simulated tactile flow provides useful
shear information in rigid-body simulation, we consider a dual-arm
object-following task adapted from~\cite{lloyd2024shear,freud2025simshear}.
A leader arm moves an object along a randomly generated 6-DoF trajectory,
while a tactile follower arm must maintain a target contact pose relative
to the moving object. Unlike prior approaches based on hand-designed
pose-and-shear tactile servoing controllers, we learn the follower
behaviour using a two-stage teacher--student RL pipeline. The task depends critically on 
shear because lateral and rotational object motions induce
tangential motion at the contact interface that cannot be observed
from contact depth alone.

\subsubsection{State, Observation, and Action Spaces}

In Stage 1, the teacher policy receives privileged state information
comprising the object pose and twist, the follower TCP pose and twist,
and low-dimensional contact information, including the contact position
and shear offset. These privileged states are used exclusively for
teacher-policy training.

In Stage~2, the student policy receives the follower TCP pose together
with the contact-depth map and simulated tactile-flow field. These
high-dimensional tactile observations are used exclusively during
student-policy training. The student policy is trained to imitate the teacher through action distillation. Since this task has smaller observation and action spaces than the dexterous in-hand manipulation task, plain action distillation
is sufficient for training the student policy. 

\subsubsection{Reward Function and Evaluation Metrics}
The reward encourages the follower TCP to maintain the target contact
pose relative to the moving object while producing stable control
behaviour. The per-step reward at time $t$ is
\[
r_t
=
-\lambda_{\mathrm{dist}} r_t^{\mathrm{dist}}
-\lambda_{\mathrm{ori}} r_t^{\mathrm{ori}}
-\lambda_{\mathrm{act}} r_t^{\mathrm{act}}\,,
\]
where the distance term \(r_t^{\mathrm{dist}}\) penalizes the Euclidean distance between the current and target relative contact positions, the orientation term \(r_t^{\mathrm{ori}}\) penalizes the cosine difference between the current and target relative orientations of the object and the follower TCP, and the action term \(r_t^{\mathrm{act}}\) penalizes excessive control actions to encourage stable follower behaviour. The coefficients $\lambda_{\mathrm{dist}}$, $\lambda_{\mathrm{ori}}$, and $\lambda_{\mathrm{act}}$ determine the relative weights of the corresponding terms.

For evaluation, we report the training curves of the position-distance reward and orientation-distance reward in simulation. For real-world experiments, we report the mean Euclidean position error and mean geodesic orientation error between the current and target relative contact poses over each episode.

\subsection{In-hand Manipulation Task Setup}\label{subsec:task_describe_in_hand}

Next, we evaluate the contributions of the rich tactile representation and PETS-Net  on the dexterous in-hand manipulation task using two 16-DoF dexterous-hand embodiments: the Allegro Hand and the LEAP Hand~\cite{shaw2023leap}. In simulation, both hands are equipped with four tactile fingertip models and use the same task formulation, tactile observation modalities, reward design, and teacher–student learning framework. Policies are trained separately for each embodiment using embodiment-specific kinematics, joint limits, action scaling, and initial grasp configurations.
The task requires the hand to maintain a stable grasp on an object while inducing continuous rotation about a desired axis. 

Our task setting is related to prior dexterous in-hand manipulation formulations~\cite{qi2023inhand,qi2024inhand,yang2024anyrotate}, but differs in both task randomization and tactile observations. Following AnyRotate~\cite{yang2024anyrotate}, we randomize the hand orientation during training and evaluation, rather than considering only a single palm-up hand orientation as in prior in-hand manipulation settings~\cite{qi2023inhand,qi2024inhand}. This makes the task more challenging because the policy must learn gravity-invariant and contact-aware manipulation behaviours across different hand poses. 
To validate the benefit of rich tactile feedback under more difficult contact conditions, we further apply unknown random external forces and torques to the object during training and evaluation. These perturbations make the manipulation problem more sensitive to contact changes, object motion, and slip-like behaviours.

\subsubsection{State, Observation and Action Spaces}

We primarily investigate the proposed high-dimensional tactile representation, where each fingertip provides simulated contact-depth and tactile-flow observations. The four fingertip observations are tiled into image-like inputs for the student policy, and we evaluate different combinations of proprioception, contact depth, tactile flow, and low-dimensional contact information.

For comparison, we also consider the following observation settings, with detailed dimensions provided in the Appendix:
\begin{enumerate}[label=(\alph*),wide=0pt]
\item \emph{Privileged State.} Used only for teacher-policy training, including object pose and motion, hand orientation, physical properties, and external disturbances.
\item \emph{Proprioception.} Following HORA~\cite{qi2023inhand}, this includes hand joint positions and current joint targets, with short-term inputs for control and longer history for student adaptation.
\item \emph{Low-dimensional Contact Information.} Following AnyRotate~\cite{yang2024anyrotate}, each fingertip provides local contact position, force magnitude, and a binary contact flag, optionally over a temporal history.
\end{enumerate}

Having defined the policy formulation above, we next introduce the experimental comparisons used to evaluate the proposed tactile modalities and policy architecture. These settings are included purely for ablation and comparison rather than as alternative components of our method.

These observation settings are used only to define the comparisons in our modality ablation experiments, where all policies use PETS-Net and differ only in their observation inputs. We compare proprioception-only policies~\cite{qi2023inhand} and existing low-dimensional contact representations~\cite{yang2024anyrotate} against our contact-depth and tactile-flow modalities, evaluated individually and in combination. Separately, we assess the contribution of PETS-Net by replacing it with a standard adaptation module under representative observation settings.

\subsubsection{Reward Function and Evaluation Metrics}

The reward encourages continuous object rotation about the desired axis while penalizing unstable object motion, deviation from the initial grasp posture, joint-target tracking error, and excessive control effort. The per-step reward at time \(t\) is
\[
r_t
=
r_t^{\mathrm{rot}}
-\lambda_{\mathrm{lin}} r_t^{\mathrm{lin}}
-\lambda_{\mathrm{pose}} r_t^{\mathrm{pose}}
-\lambda_{\mathrm{track}} r_t^{\mathrm{track}}
-\lambda_{\mathrm{work}} r_t^{\mathrm{work}}\,,
\]
where \(r_{\mathrm{rot},t}\) rewards the object angular velocity along the desired rotation axis, while \(r_t^{\mathrm{lin}}\), \(r_t^{\mathrm{pose}}\), \(r_t^{\mathrm{track}}\), and \(r_t^{\mathrm{work}}\) penalize object translation, deviation from initial grasp posture, joint-target tracking error, and control effort, respectively. The coefficients \(\lambda_{\mathrm{lin}}\), \(\lambda_{\mathrm{pose}}\), \(\lambda_{\mathrm{track}}\), and \(\lambda_{\mathrm{work}}\) determine the relative weights of the corresponding penalty terms. 

For evaluation, we report metrics that measure object retention and rotation performance~\cite{qi2023inhand,yang2024anyrotate}. Specifically, \textit{Rotation Count} (Rot.) measures the accumulated object rotation about the commanded target axis, expressed as the equivalent number of complete revolutions. For simulation experiments, we report Rot. together with \textit{Time-to-Fail} (TTF) and \textit{Fall Rate}. TTF measures the elapsed simulation time before the object is dropped, while Fall Rate denotes the percentage of evaluation episodes in which the object is dropped before the maximum episode duration. For real-world experiments, each trial lasts for at most \(600\) control steps, corresponding to \(30\,\mathrm{s}\) of execution. We report Rot. together with \textit{Time to Termination} (TTT), defined as the elapsed time before the object is dropped, becomes stuck, or the trial otherwise becomes invalid. A trial that remains valid for the full evaluation duration achieves the maximum TTT of \(30\,\mathrm{s}\).


\section{Results and Analysis}

\subsection{Simulation Results: Ablation Study of Tactile Modalities}

\begin{figure*}[t]
\centering

\begin{subfigure}[t]{0.31\textwidth}
    \centering
    \refstepcounter{subfigure}
    \label{fig:abl_main}
    \begin{minipage}[t]{0.07\linewidth}
        \vspace{0.42\linewidth}
        \centering
        (\alph{subfigure})
    \end{minipage}%
    \hspace{0.01\linewidth}%
    \begin{minipage}[t]{0.92\linewidth}
        \vspace{0pt}
        \includegraphics[width=\linewidth]{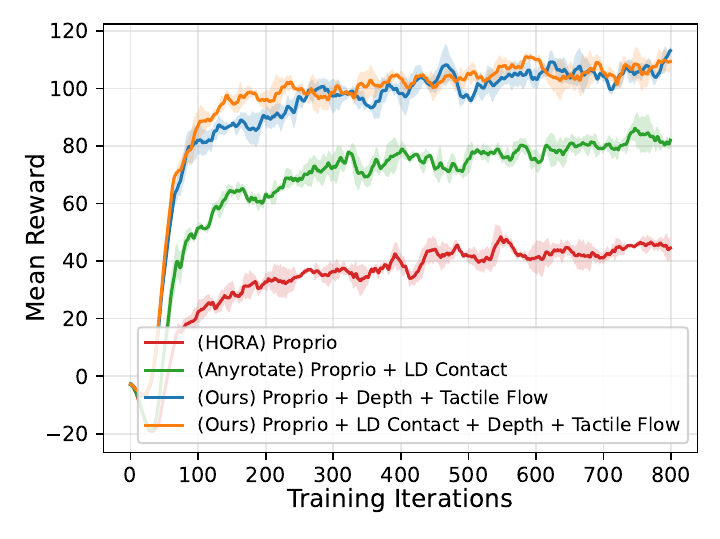}
    \end{minipage}
\end{subfigure}
\hfill
\begin{subfigure}[t]{0.31\textwidth}
    \centering
    \refstepcounter{subfigure}
    \label{fig:abla_rich_tactile}
    \begin{minipage}[t]{0.07\linewidth}
        \vspace{0.42\linewidth}
        \centering
        (\alph{subfigure})
    \end{minipage}%
    \hspace{0.01\linewidth}%
    \begin{minipage}[t]{0.92\linewidth}
        \vspace{0pt}
        \includegraphics[width=\linewidth]{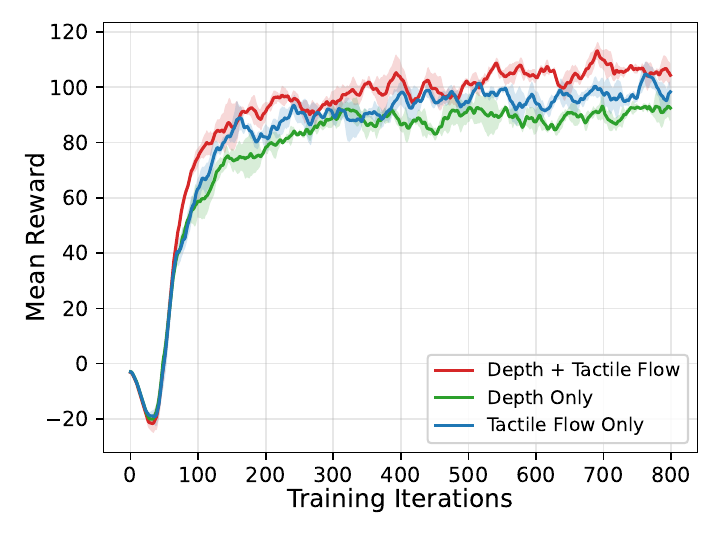}
    \end{minipage}
\end{subfigure}
\hfill
\begin{subfigure}[t]{0.34\textwidth}
    \centering
    \refstepcounter{subfigure}
    \label{fig:abla_hist_len_and_diff_modality_comparison}
    \begin{minipage}[t]{0.07\linewidth}
        \vspace{0.42\linewidth}
        \centering
        (\alph{subfigure})
    \end{minipage}%
    \hspace{0.01\linewidth}%
    \begin{minipage}[t]{0.92\linewidth}
        \vspace{0pt}
        \includegraphics[
            width=\linewidth,
            trim={0mm 4mm 0mm 6mm},
            clip
        ]{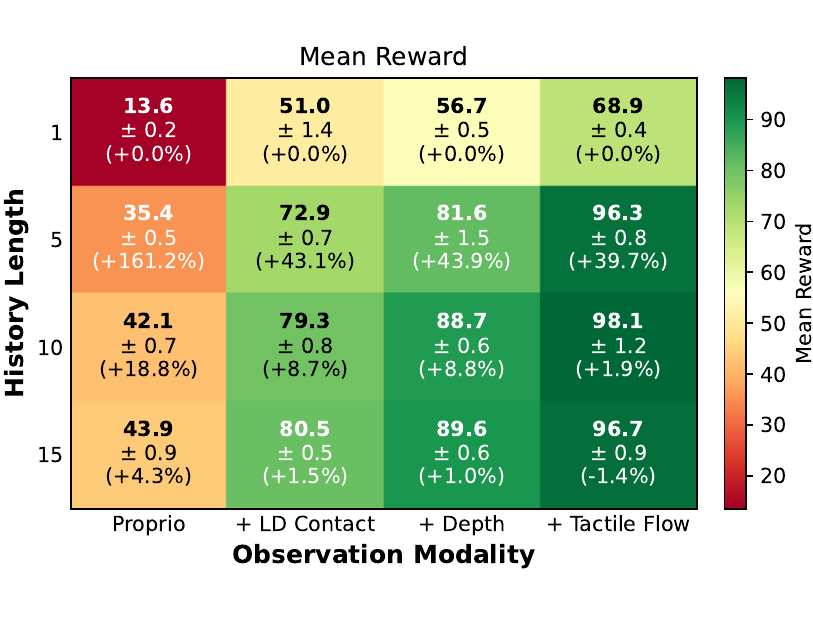}
    \end{minipage}
\end{subfigure}

\vspace{-0.5em}

\caption{\small Ablation studies of tactile observation richness and temporal history in the in-hand manipulation task.
(a) Policy input modalities compared with proprioception-only HORA~\cite{qi2023inhand} and LD contact AnyRotate~\cite{yang2024anyrotate}.
(b) Comparison of contact depth, tactile flow, and their combination.
(c) Final reward across modalities and history lengths, with percentages showing gains over the preceding shorter history. Dense tactile observations, particularly tactile flow, approach saturation with shorter histories. Solid curves show the mean over three random seeds, with shaded regions denoting ±1 standard deviation. Values in (c) report mean ± standard deviation across three random seeds.}
\label{fig:abla_obs_hist}
\end{figure*}
\subsubsection{Training Performance across Tactile Modalities}

We compare our rich, high-dimensional tactile modalities with two baselines:  a proprioception-only setting, corresponding to the HORA-style baseline~\cite{qi2023inhand}, as well as proprioception plus low-dimensional contact information, corresponding to the AnyRotate-style tactile baseline~\cite{yang2024anyrotate}. Experiments show that richer tactile observations substantially improve both convergence speed and final policy performance (Fig.~\ref{fig:abl_main}). 

The proprioception-only policy~\cite{qi2023inhand} learns slowly and converges to a much lower reward, indicating that proprioceptive history alone provides limited information about the local fingertip-object interaction. Adding low-dimensional contact information (Anyrotate~\cite{yang2024anyrotate}) improves training, as compact contact position and force cues provide useful information about contact state, which is consistent with the results reported in~\cite{yang2024anyrotate}. However, this representation still loses detailed spatial contact geometry and tangential motion information. In contrast, our rich, high-dimensional tactile observations, including contact depth and tactile flow, lead to faster convergence and higher final reward. Notably, adding low-dimensional contact information on top of the high-dimensional tactile modalities gives only marginal additional improvement, suggesting that the proposed dense tactile observations already capture the key contact information needed for policy learning.

\subsubsection{Contribution of Contact Depth and Tactile Flow}
\begin{figure}
  \centering
  \includegraphics[width=1\linewidth]{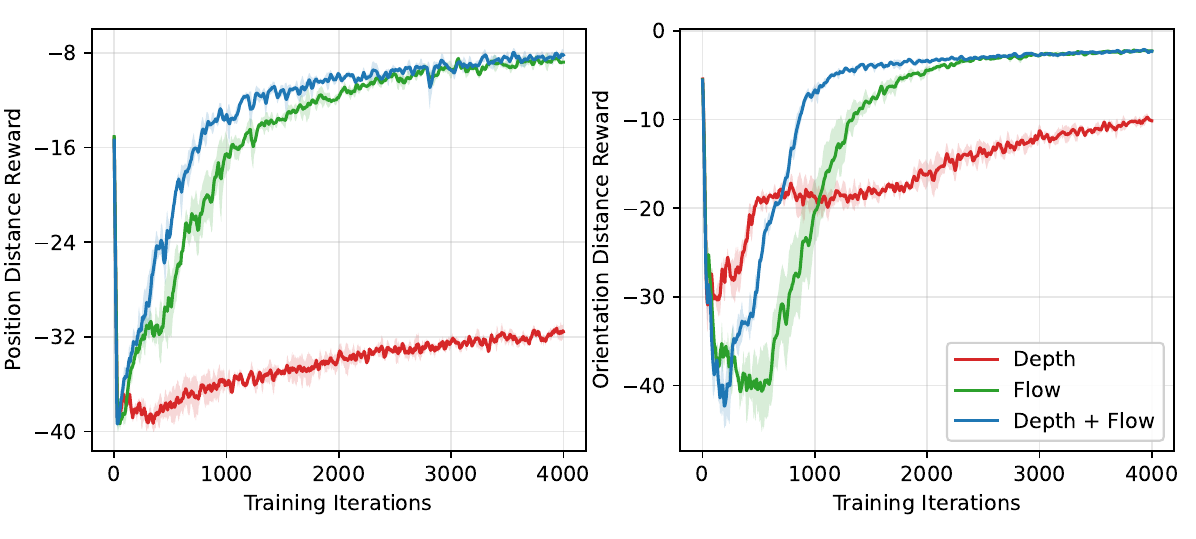}
  \vspace{-0.5em}
  \caption{\small Ablation studies of contact depth and tactile flow for the shear-sensitive dual-arm object-following task. Solid curves show the mean over three random seeds, with shaded regions denoting ±1 standard deviation.}
  \label{fig:abla_ts}
\end{figure}

To assess the contributions of contact depth and tactile flow, we first consider the shear-sensitive dual-arm object-following task. Fig.~\ref{fig:abla_ts} shows that tactile flow outperforms depth alone. For approximately flat contacts, depth maps mainly capture normal deformation and provide limited information about tangential motion. Tactile flow directly encodes shear displacement and implicitly reflects contact area, improving both translational and rotational tracking. Combining depth and flow accelerates learning but achieves similar final performance to flow alone, indicating that flow is the dominant modality.

We next examine whether this trend extends to the more complex in-hand manipulation task. As shown in Fig.~\ref{fig:abla_rich_tactile}, tactile flow outperforms contact depth alone, highlighting the importance of tangential-motion cues for object rotation. Combining both modalities achieves the fastest convergence and highest final reward: depth captures contact geometry, complementing flow that describes fingertip--object shear motion. This advantage is further reflected in the simulated task evaluation (Table~\ref{tab:simulation_results}), where combining depth and flow achieves the longest time-to-fail and lowest fall rate both within the training distribution ($27.01,\mathrm{s}$, $11.76\%$) and on OOD objects ($25.08,\mathrm{s}$, $18.69\%$), outperforming either modality alone.

Overall, both tasks demonstrate that tactile flow provides the primary shear-sensitive information needed for contact-rich control, consistently outperforming contact depth alone. Contact depth remains complementary by providing normal contact geometry and distribution cues. 

\begin{table*}[t]
\centering
\caption{In-hand manipulation performance across tactile observation modalities on within-distribution and out-of-distribution objects in simulation. Results are reported as mean $\pm$ standard deviation.}
\label{tab:simulation_results}
\resizebox{\textwidth}{!}{
\begin{tabular}{l ccc ccc}
\toprule
\multirow{2}{*}{\textbf{Observations}} &
\multicolumn{3}{c}{\textbf{Within Training Distribution}} &
\multicolumn{3}{c}{\textbf{Out-of-Distribution}} \\
\cmidrule(lr){2-4}\cmidrule(lr){5-7}
& \textbf{Time-to-Fail (s)} & \textbf{Rot. Counts} &  \textbf{Fall Rate (\%)}
& \textbf{Time-to-Fail (s)} & \textbf{Rot. Counts} & \textbf{Fall Rate (\%)} \\
\midrule
Oracle Teacher & $27.45\pm2.13$ & $1.88\pm0.66$ & 9.37 & $25.11\pm3.51$ & $1.74\pm0.81$ & 18.09 \\
\midrule
Prop. & $19.87\pm4.04$ & $0.97\pm0.88$ & 47.66 & $16.64\pm4.24$ & $0.88\pm0.85$ & 57.85 \\
Prop. + LD Contact& $25.18\pm3.34$ & $1.21\pm0.90$ & 21.33 & $22.00\pm4.09$ & $1.19\pm0.88$ & 31.21 \\
Prop. + Flow (Ours)  & $25.57\pm3.30$ & $1.24\pm0.84$ & 17.45 & $22.80\pm3.89$ & $1.19\pm0.91$ & 29.82 \\
Prop. + Depth (Ours)& $26.19\pm3.14$ & $1.21\pm0.89$ & 14.26 & $22.27\pm4.08$ & $1.20\pm0.91$ & 29.91 \\
Prop. + Depth + Flow (Ours) &
$\mathbf{27.01\pm2.82}$ &
$\mathbf{1.28\pm0.85}$ &
\textbf{11.76} &
$\mathbf{25.08\pm3.50}$ &
$\mathbf{1.27\pm0.94}$ &
\textbf{18.69} \\
\bottomrule
\end{tabular}}
\end{table*}

\subsubsection{Rich Tactile Observations Reduce the Required Temporal Context}

We now compare the history lengths within each modality and modalities under fixed history lengths (Fig.~\ref{fig:abla_hist_len_and_diff_modality_comparison}).

For proprioception-only and low-dimensional contact inputs, longer histories substantially improve learning speed, final reward, and convergence stability. Instantaneous observations provide insufficient information about contact transitions, object motion, and previous action effects, while longer histories allow the policy to reconstruct these interaction dynamics. Performance generally improves up to a history length of approximately 10--15 steps before approaching saturation.

In contrast, policies using contact-depth or tactile-flow observations reach strong performance for only 5--10 history steps. These modalities make the interaction state more directly observable: contact depth describes contact depth and geometry, while tactile flow explicitly captures local shear motion. Their combination therefore provides complementary normal and tangential contact information.

Overall, the results reveal a trade-off between observation richness and temporal context. Sparse or indirect observations require longer histories, whereas dense tactile feedback reduces the memory burden of the temporal adaptor while improving policy performance.

\subsubsection{Emergent Contact-Aware Pinch-Gaiting Behaviour}

\begin{figure}
  \centering
  \includegraphics[width=1\linewidth]{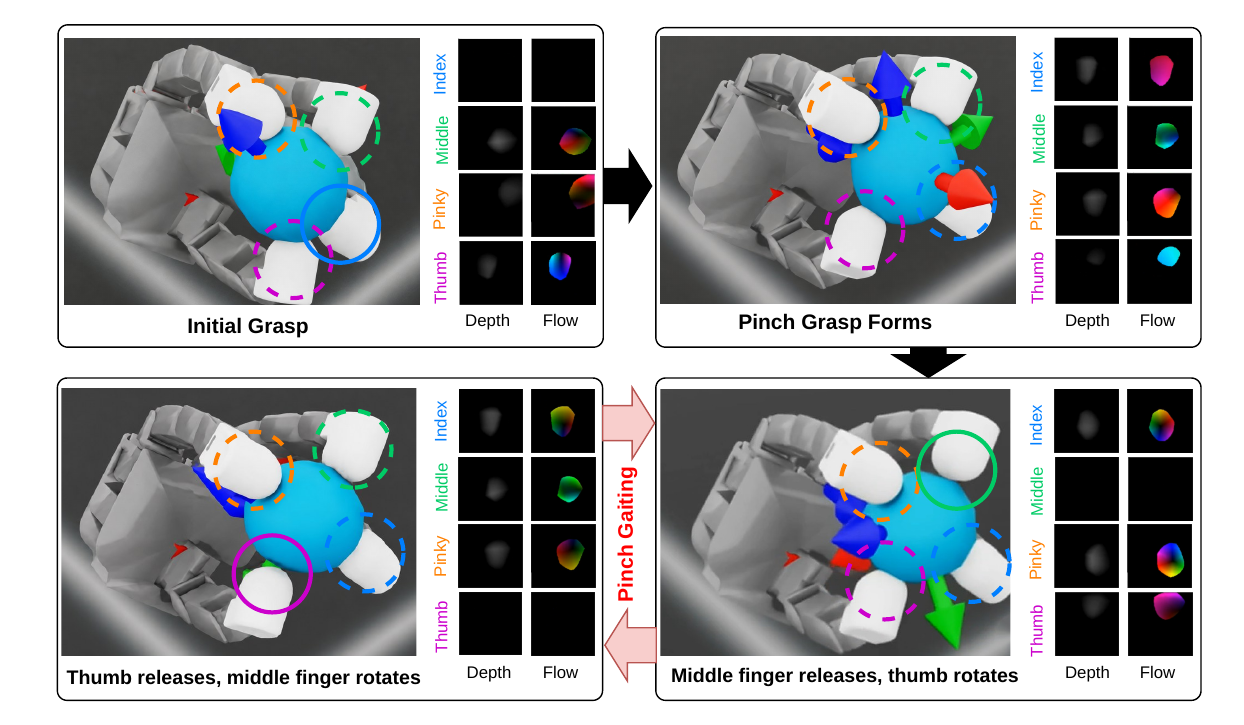}
  \vspace{-1em}
  \caption{\small Emergent Contact-Aware Pinch-Gaiting Behaviour with Dense Contact-Depth and Tactile-Flow Feedback. Dotted circles denote in-contact while solid ones denote no contact. }
  \label{fig:pinch_gaiting}
\end{figure}

As shown in Fig.~\ref{fig:pinch_gaiting}, policies trained with rich tactile observations develop a new emergent pinch-gaiting strategy. Two opposing fingers stabilize the object, while the remaining fingers alternately make and break contact to generate rotation. The depth maps capture these contact transitions, whereas the coherent tactile-flow patterns indicate consistent tangential motion during active contact. This structured behaviour was not observed with proprioception-only or low-dimensional contact inputs, suggesting that dense tactile feedback supports coordinated grasp stabilization and fingertip gaiting.

\subsection{Simulation Results: Ablation Study of PETS-Net}

\subsubsection{Positional Encoding Applied to the Proprioceptive-Action History}

\begin{figure*}[b]
\centering

\begin{subfigure}[t]{0.32\textwidth}
    \centering
    \refstepcounter{subfigure}
    \label{fig:abla_pe}
    \begin{minipage}[t]{0.07\linewidth}
        \vspace{0.42\linewidth}
        \centering
        (\alph{subfigure})
    \end{minipage}%
    \hspace{0.01\linewidth}%
    \begin{minipage}[t]{0.92\linewidth}
        \vspace{0pt}
        \includegraphics[width=\linewidth]{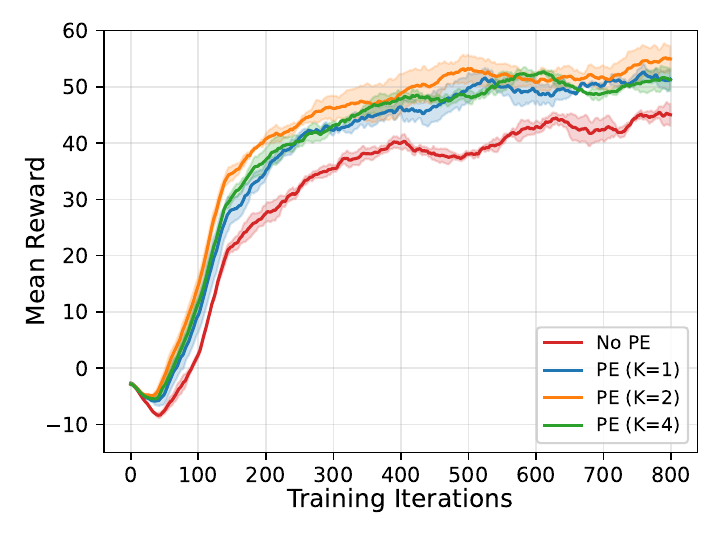}
    \end{minipage}
\end{subfigure}
\hfill
\begin{subfigure}[t]{0.32\textwidth}
    \centering
    \refstepcounter{subfigure}
    \label{fig:abla_temporal_component}
    \begin{minipage}[t]{0.07\linewidth}
        \vspace{0.42\linewidth}
        \centering
        (\alph{subfigure})
    \end{minipage}%
    \hspace{0.01\linewidth}%
    \begin{minipage}[t]{0.92\linewidth}
        \vspace{0pt}
        \includegraphics[width=\linewidth]{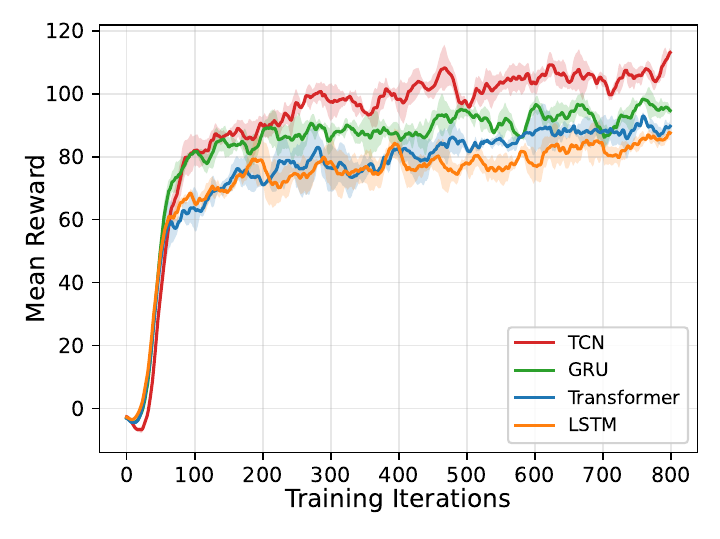}
    \end{minipage}
\end{subfigure}
\hfill
\begin{subfigure}[t]{0.32\textwidth}
    \centering
    \refstepcounter{subfigure}
    \label{fig:abla_spatial_temporal}
    \begin{minipage}[t]{0.07\linewidth}
        \vspace{0.42\linewidth}
        \centering
        (\alph{subfigure})
    \end{minipage}%
    \hspace{0.01\linewidth}%
    \begin{minipage}[t]{0.92\linewidth}
        \vspace{0pt}
        \includegraphics[width=\linewidth]{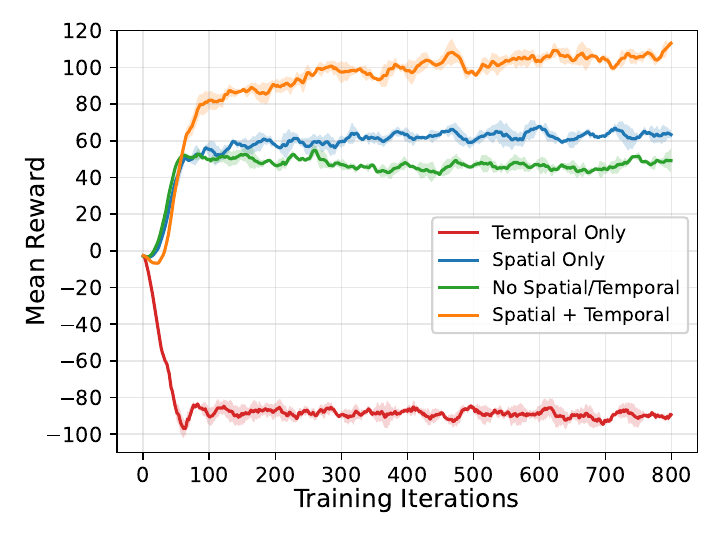}
    \end{minipage}
\end{subfigure}

\vspace{-0.3em}

\caption{\small Ablation studies of PETS-Net.
(a) Effect of positional encoding frequency in the adaptor.
(b) Effect of temporal-history encoder structure.
(c) Effect of spatial and temporal components. Solid curves show the mean over three random seeds, with shaded regions denoting ±1 standard deviation.}
\label{fig:abla_petsnet}

\end{figure*}

We first study the effect of applying positional encoding (PE) to the proprioceptive-action history. As shown in Fig.~\ref{fig:abla_pe}, all PE-based variants outperform the model without PE. The no-PE variant converges more slowly and reaches a lower final reward, indicating that raw proprioceptive-action histories alone are less effective for representing the temporal evolution of hand-object interaction. In contrast, PE-based variants achieve faster reward growth and higher final performance, with dependence on the number of frequency components: $K=2$ is best with $K=1,4$ competitive. Increasing $K$ further does not provide additional benefit, which because higher-frequency encodings increase the feature dimensionality and make the adaptor harder to optimize. These results suggest that a moderate-frequency positional encoding provides useful temporal structure for the adaptor.
\subsubsection{Temporal Encoder Ablation}

Second, we compare several temporal encoders, including a Temporal Convolutional Network (TCN)~\cite{bai2018empirical}, a Gated Recurrent Unit (GRU)~\cite{cho2014learning}, a Long Short-Term Memory (LSTM) network~\cite{hochreiter1997long}, and a Transformer~\cite{vaswani2017attention}, while keeping the remaining PETS-Net components unchanged. As shown in Fig.~\ref{fig:abla_temporal_component}, the TCN achieves the highest final reward and stable convergence, while the GRU provides comparable but slightly lower performance.

The TCN also provides a favorable memory--performance trade-off in our setting. Policy performance largely saturates with a 10-frame history, for which lightweight temporal convolutions are sufficient to aggregate the short-term contact dynamics encoded by the CNN. The Transformer does not improve performance under this history horizon and requires substantially more VRAM, preventing evaluation with 15- or 20-frame histories on our hardware (specified in Appendix). These results support the use of the TCN as an efficient temporal adaptor for the current tactile representation and training setting, rather than indicating a general advantage over attention-based or recurrent architectures.

\subsubsection{Ablation on Spatial and Temporal Components}

Third, we ablate the spatial and temporal components of PETS-Net using contact-depth and tactile-flow observations. The full model combines CNN-based spatial encoding with temporal-history modeling. We compare it against three reduced variants: temporal history without the spatial encoder, single-step input with spatial encoding, and a single-step baseline without either component. As shown in Fig.~\ref{fig:abla_spatial_temporal}, the full spatio-temporal model achieves the highest reward. Spatial encoding alone improves over the single-step baseline, confirming that CNN features capture useful contact geometry and deformation. However, its lower performance indicates that a single tactile observation cannot fully represent contact evolution. Conversely, the temporal-history-only variant performs poorly, suggesting that temporal modeling is ineffective without first extracting meaningful spatial features from the high-dimensional tactile maps. Overall, the results support the PETS-Net design. CNN-based encoding extracts spatial contact information, while the TCN captures temporal changes in contact and object motion. Combining both components enables more effective policy distillation and achieves faster convergence and higher final reward than the reduced variants.

\subsection{Real-to-Sim Tactile Grounding Evaluation}
\label{subsec:Real-to-Sim_flow_transfer}

\begin{table*}[b]
\centering
\caption{\small Tactile-flow-based shear estimation performance for simulated, real, and real-to-sim inputs. Each entry reports MAE\(/R^2\), averaged over 8 independently trained regressors (identical data, stochastic weight initialization/batch order). Sim denotes the in-domain result obtained directly from simulated test flow, Real denotes the result obtained directly from real tactile image pairs, and Real-to-Sim denotes the result obtained from real tactile image pairs translated into simulated tactile flow before shear estimation.}
\label{tab:Real-to-Sim_flow_transfer}
\resizebox{\textwidth}{!}{
\begin{tabular}{llcccccc}
\toprule
Sensor & Press depth
& \multicolumn{3}{c}{\(\bar v_{xy}\) (\(\mathrm{mm/s}\))}
& \multicolumn{3}{c}{\(\omega_z\) (\(^{\circ}/\mathrm{s}\))} \\
\cmidrule(lr){3-5}
\cmidrule(lr){6-8}
& & Sim & Real & Real-to-Sim & Sim & Real & Real-to-Sim \\
\midrule
GelSight Mini, markerless & 1\,mm & 0.30/0.99 & 1.79/-0.02 & 0.34/0.98 & 1.33/0.99 & 4.50/0.89 & 1.18/0.99 \\
DIGIT, markerless         & 1\,mm & 0.29/0.99 & 0.64/0.94  & 0.48/0.96 & 1.32/0.99 & 3.61/0.92 & 3.33/0.92 \\
9DTact, markerless        & 1\,mm & 0.30/0.99 & 0.45/0.97  & 0.35/0.98 & 1.29/0.99 & 2.02/0.97 & 1.29/0.99 \\
GelSight Mini, markers    & 1\,mm & 0.30/0.99 & 0.35/0.98  & 0.27/0.99 & 1.30/0.99 & 0.90/1.00 & 0.97/1.00 \\
Flat TacTip               & 1\,mm & 0.30/0.99 & 0.26/0.99  & 0.24/0.99 & 1.30/0.99 & 1.02/0.99 & 1.00/0.99 \\
Hemispherical TacTip      & 1\,mm & 0.30/0.99 & 0.27/0.99  & 0.29/0.99 & 1.42/0.99 & 2.09/0.98 & 2.22/0.97 \\
Hemispherical TacTip      & 5\,mm & 0.29/0.99 & 0.25/0.99  & 0.27/0.99 & 1.32/0.99 & 0.91/1.00 & 1.25/0.99 \\
\bottomrule
\end{tabular}
}
\vspace{2pt}

{\footnotesize $^*$Velocities are derived from per-frame shear displacement MAE assuming 25\,Hz sensor sample rate. \(\bar v_{xy}\) values average the per-axis \(\bar v_{x}\) and \(\bar v_{y}\) values.}
\end{table*}

\begin{figure}[t]
  \centering
  \includegraphics[width=1\linewidth]{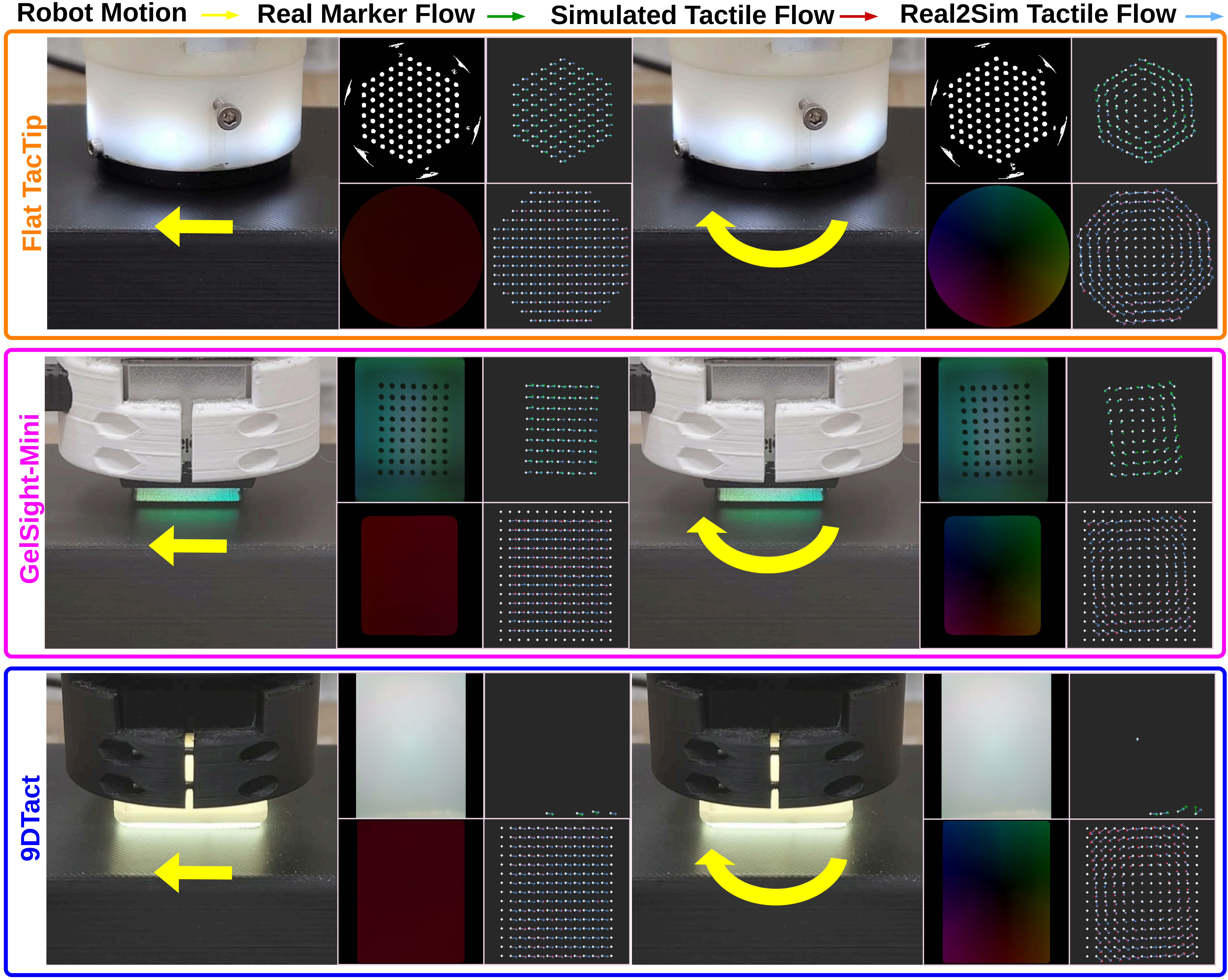}
    \caption{\small Test-time demonstration of tactile real-to-sim transfer for representative sliding and twisting motions using flat tactile sensors: Flat TacTip, 9DTact, and GelSight Mini withmarkers. In each example, the large image on the left shows the real-world sensor setup, while the panels on the right show the real tactile images, real vs. generated sparse flow fields, the generated pixel-level dense flow field, and simulated vs. generated sparse flow fields. The results demonstrate real-to-sim tactile-flow transfer across sensors with different sensing principles.
}
  \label{fig:gan_gs_flat_tactip_output_tactile_flow}
\end{figure}

We now conduct real-to-sim experiments to validate whether the proposed simulated tactile representations can be grounded in real-world sensor observations through the proposed tactile transfer pipeline. Specifically, we evaluate whether the real-to-sim translators can recover simulated tactile-flow representations from real observations of various sensors while preserving shear information for downstream motion estimation. The evaluation covers various camera-based tactile sensors: a markerless GelSight Mini~\cite{yuan2017gelsight}, DIGIT~\cite{lambeta2020digit}, 9DTact~\cite{lin20249dtact}, a GelSight Mini with markers, a flat TacTip~\cite{church2022tactile}, and a hemispherical TacTip~\cite{lin2022tactilegym2} (Fig.~\ref{fig:overview}; contact and shear pose ranges given in Appendix C. 

For each sensor, we train FlowMotionNet $f_{\mathrm{FMN}}$, a CNN regressor that maps simulated tactile flow to planar motion $(v_x,v_y,\omega_z)$. The regressor is trained exclusively on simulated data and then frozen. It is evaluated on 1) held-out simulated tactile flow; and 2) on unseen real tactile-image pairs translated by the corresponding real-to-sim model:
\[
\hat{\mathbf{F}}
=
G(\mathbf{X}^{\mathrm{real}}),
\qquad
(\hat{v}_x,\hat{v}_y,\hat{\omega}_z)
=
f_{\mathrm{FMN}}(\hat{\mathbf{F}}).
\]
The simulated results provide an in-domain reference, while evaluation on translated real inputs measures the performance of the full real-to-sim transfer pipeline. For comparison, we also train and evaluate a separate regressor directly on real tactile image pairs to estimate the same planar motion components. Comparing direct image-based estimation with estimation from the generated simulated tactile flow allows us to assess how introducing the intermediate flow representation affects shear-estimation accuracy.

Table~\ref{tab:Real-to-Sim_flow_transfer} reports the mean absolute error (MAE) and coefficient of determination ($R^2$) for simulated, real, and real-to-sim inputs, averaged over eight independently trained regressors. Translational results are averaged across the two planar axes. All sensor configurations achieve similar performance on simulated inputs, with translational MAEs of $0.29$--$0.30\,\mathrm{mm/s}$, rotational MAEs of $1.29$--$1.42^\circ/\mathrm{s}$, and $R^2=0.99$.

Sensors with explicit markers provide accurate shear estimates directly from real inputs, with relatively small changes following real-to-sim translation. At a $1\,\mathrm{mm}$ press depth, the GelSight Mini with markers and flat TacTip achieve real-to-sim translational MAEs of $0.27$ and $0.24\,\mathrm{mm/s}$ and rotational MAEs of $0.97^\circ/\mathrm{s}$ and $1.00^\circ/\mathrm{s}$, respectively, with $R^2\geq0.99$. 

In contrast, markerless sensors benefit more substantially from real-to-sim translation. For the markerless GelSight Mini, translation reduces the translational MAE from $1.79$ to $0.34\,\mathrm{mm/s}$ and the rotational MAE from $4.50^\circ/\mathrm{s}$ to $1.18^\circ/\mathrm{s}$, while increasing the corresponding $R^2$ values from $-0.02$ to $0.98$ and from $0.89$ to $0.99$. Similarly, the markerless 9DTact improves from $0.45$ to $0.35\,\mathrm{mm/s}$ and from $2.02^\circ/\mathrm{s}$ to $1.29^\circ/\mathrm{s}$. The markerless DIGIT also improves, although its rotational MAE remains higher at $3.33^\circ/\mathrm{s}$, with $R^2=0.92$. These results indicate that real-to-sim translation can mitigate the challenges of shear estimation from markerless tactile images, although its effectiveness varies across sensors.

The hemispherical TacTip benefits from deeper contact. Increasing the press depth from $1$ to $5\,\mathrm{mm}$ maintains similar real-to-sim translational accuracy, with MAEs of $0.29$ and $0.27\,\mathrm{mm/s}$, while reducing the rotational MAE from $2.22^\circ/\mathrm{s}$ to $1.25^\circ/\mathrm{s}$ and increasing rotational $R^2$ from $0.97$ to $0.99$. The same trend is observed for direct real inputs, where the rotational MAE decreases from $2.09^\circ/\mathrm{s}$ to $0.91^\circ/\mathrm{s}$. This improvement is consistent with deeper contact making rotational shear more observable through larger marker displacements, exploiting the TacTip's soft hemispherical elastomer and large deformation range.

Fig.~\ref{fig:gan_gs_flat_tactip_output_tactile_flow} shows qualitatively similar real-to-sim tactile-flow transfer across the flat TacTip, 9DTact, and GelSight Mini with markers.
Although the GelSight Mini achieves slightly better transfer accuracy because its higher surface friction suppresses slip during data collection, we select the hemispherical TacTip for downstream in-hand manipulation and object-following tasks. Its greater durability and larger elastic deformation range make it better suited to sustained contact-rich interaction, whereas the GelSight Mini is more fragile and constrained in deformation range~\cite{lin2022tactilegym2}.

\subsection{Real-world Evaluation}

\begin{figure}
\centering
\setlength{\tabcolsep}{0pt}
\renewcommand{\arraystretch}{0.1}

\begin{tabular}{@{}ccc@{}}
    \includegraphics[width=0.28\columnwidth]{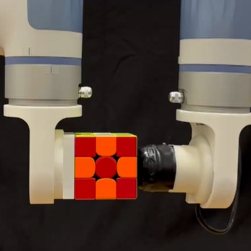} &
    \includegraphics[width=0.28\columnwidth]{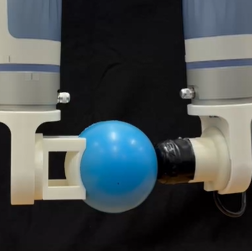} &
    \includegraphics[width=0.28\columnwidth]{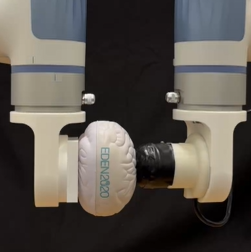} \\[1.0em]

    \includegraphics[width=0.333\columnwidth,trim={8mm 1mm 1mm 8mm},clip]{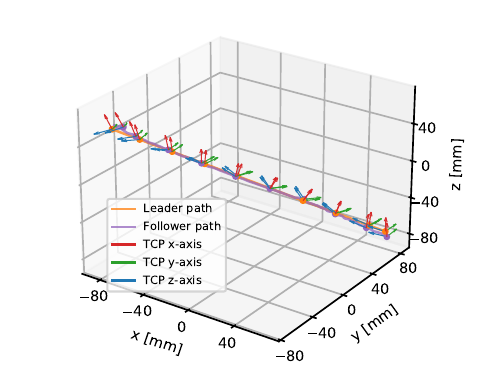} &
    \includegraphics[width=0.333\columnwidth,trim={8mm 1mm 1mm 8mm},clip]{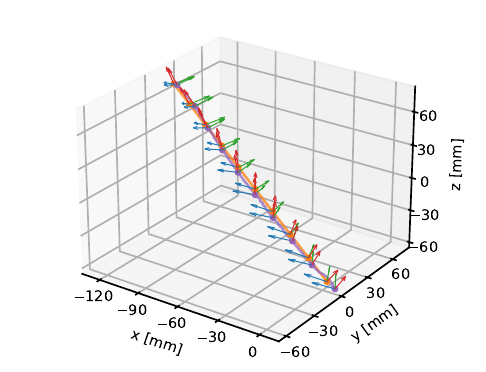} &
    \includegraphics[width=0.333\columnwidth,trim={8mm 1mm 1mm 8mm},clip]{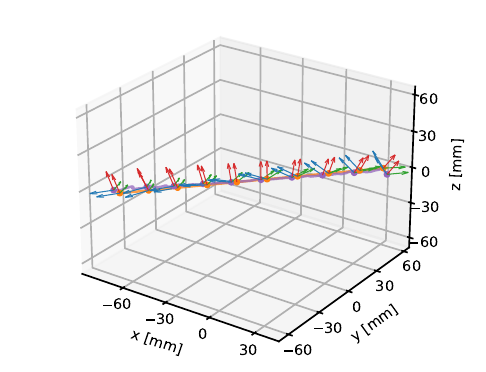} \\

    \includegraphics[width=0.27\columnwidth,trim={7mm 1mm 1mm 8mm},clip]{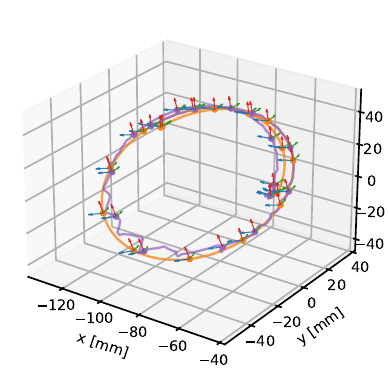} &
    \includegraphics[width=0.333\columnwidth,trim={8mm 1mm 1mm 8mm},clip]{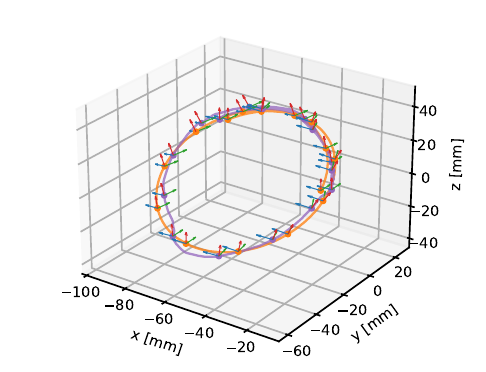} &
    \includegraphics[width=0.333\columnwidth,trim={8mm 1mm 1mm 8mm},clip]{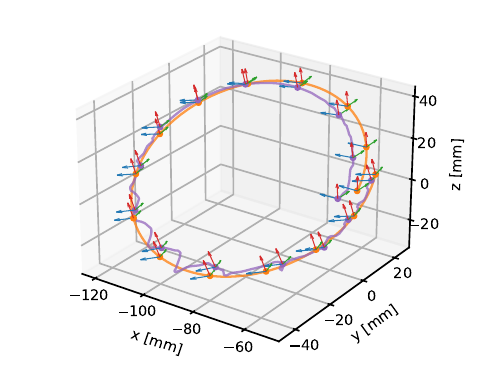}
\end{tabular}

\captionof{figure}{\small Real-world object-following trajectories. Top row: rigid cube, soft ball, and foam brain in the experiments. Orange and purple curves denote the leader and follower paths, respectively; red, green, and blue arrows denote the $x$-, $y$-, and $z$-axes of both TCPs' frames. Second and third rows: straight-line and circular motions.}
\label{fig:obj_follow_real_3x2}
\centering
\captionof{table}{\small Mean position and orientation errors in the real-world dual-arm object-following task.}
\label{tab:object_trajectory_error}
\vspace{1em}
\scriptsize
\setlength{\tabcolsep}{3pt}
\renewcommand{\arraystretch}{0.9}
\begin{tabular}{@{}llcc@{}}
\hline
\textbf{Object} & \textbf{Trajectory} & \textbf{Pos. Error (mm)} & \textbf{Orient. Error (rad)} \\
\hline
\multirow{2}{*}{Rigid cube}
& Straight & $0.91$ & $0.019$ \\
& Circle   & $1.42$ & $0.021$ \\
\hline
\multirow{2}{*}{Soft ball}

& Straight & $1.75$ & $0.036$ \\
& Circle  & $2.87$ & $0.054$ \\
\hline
Foam brain
& Straight & $4.20$ & $0.041$ \\
& Circle   & $6.43$ & $0.057$ \\
\hline
\end{tabular}
\label{tab:obj_follow_error}
\vspace{-0.5em}
\end{figure}

\subsubsection{Dual-arm Object-following}

We evaluate the learned policy on the real dual-arm object-following task using three objects (Fig.~\ref{fig:obj_follow_real_3x2} top row) and two motion patterns including straight-line and circular motions. 
Notably, the policy is trained only in simulation using a rigid cube under randomized straight-line motions, while the real-world evaluation includes unseen circular motions and objects with different geometry, compliance, and surface properties. As shown in Table~\ref{tab:obj_follow_error} and Fig.~\ref{fig:obj_follow_real_3x2} (second and third rows), the follower arm consistently tracks the leader motion and maintains stable tactile contact across all tested settings. The best performance is achieved on the rigid cube, which is closest to the simulated training condition (mean position/orientation errors 0.91~mm/0.019~rad and 1.42~mm/0.021~rad for straight-line and circular motion). For the soft ball, the errors increase moderately (1.75~mm/0.036~rad and 2.87~mm/0.054~rad), reflecting the additional difficulty introduced by object compliance. The foam brain is the most challenging object due to its irregular surface texture and more complex contact geometry, resulting in larger errors (4.20~mm/0.041~rad and 6.43~mm/0.057~rad). Nevertheless, the policy remains effective under these out-of-distribution conditions, with all mean position errors below 6.5~mm and orientation errors below 0.06~rad. These results demonstrate that the tactile-flow-based policy trained in simulation can transfer to the real robot and generalize beyond the training distribution in both motion type and object properties.

\subsubsection{In-hand Manipulation}

\begin{figure}
  \centering
  \includegraphics[width=1\linewidth]{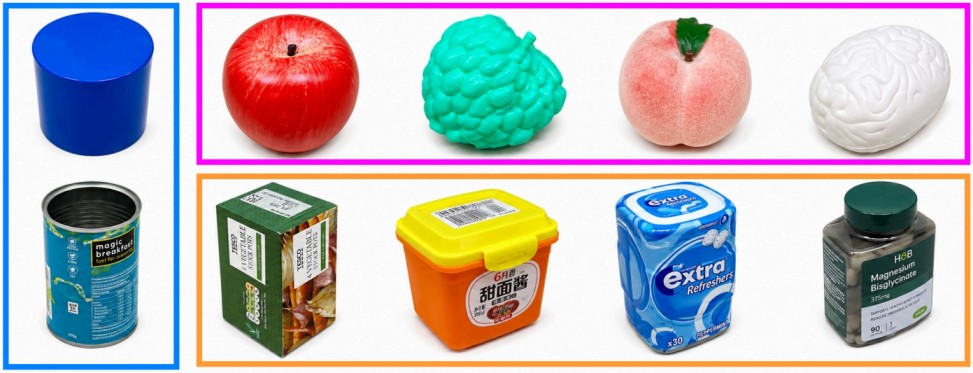}
  \vspace{-0.5em}
  \caption{\small Three sets of OOD objects used for real-world in-hand manipulation evaluation. Cylinder-shape objects (blue box): blue cylinder toy, cylindrical can. Round-shape objects: apple, sweetsop, peach, and brain-shaped foam toy. Cube-shape objects: small paper box, sauce container, chewing-gum container, and supplement bottle. The objects exhibit substantial variations in geometry, size, mass distribution, surface texture, friction, and compliance.}
  \label{fig:hand_object_set}
\end{figure}
\begin{figure*}
    \centering
    \setlength{\tabcolsep}{2pt}
    \begin{tabular}{@{}cc@{}}
        \includegraphics[width=0.49\textwidth]{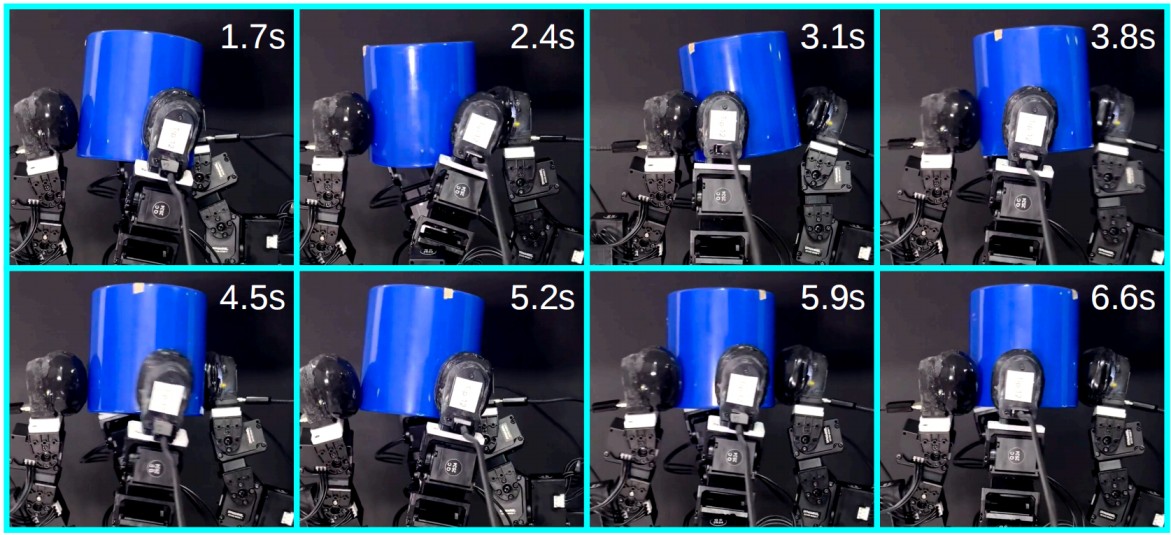} &
        \includegraphics[width=0.49\textwidth]{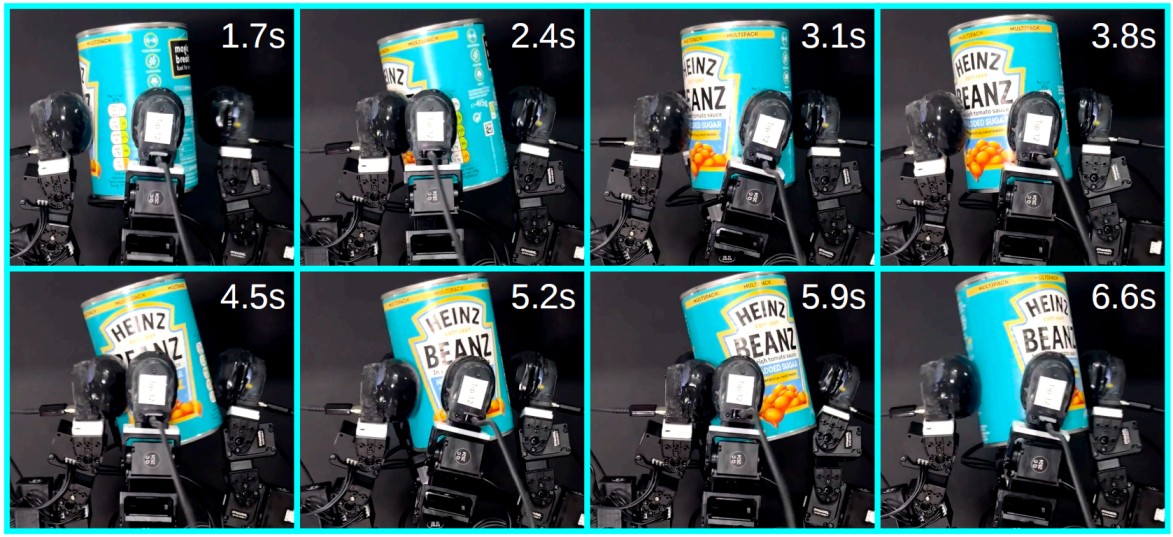} \\[-0.3em]
        \includegraphics[width=0.49\textwidth]{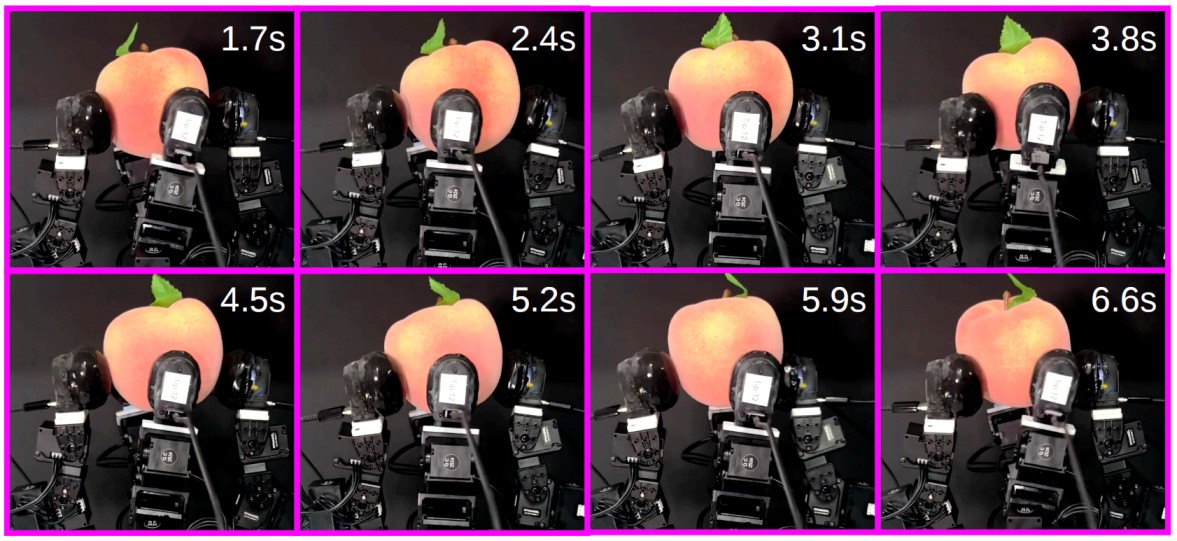} &
        \includegraphics[width=0.49\textwidth]{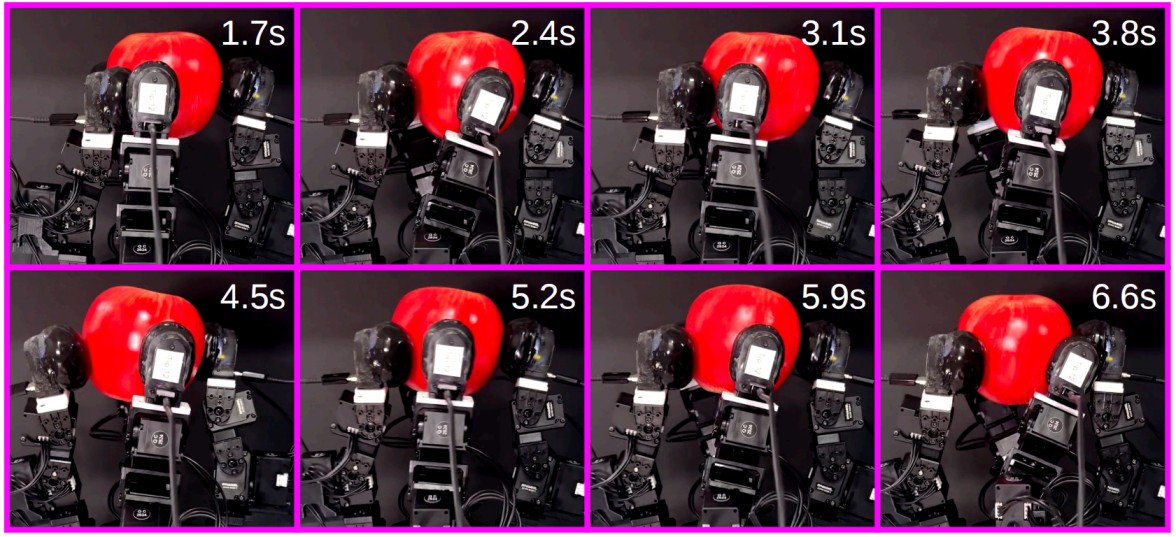} \\[-0.3em]
        \includegraphics[width=0.49\textwidth]{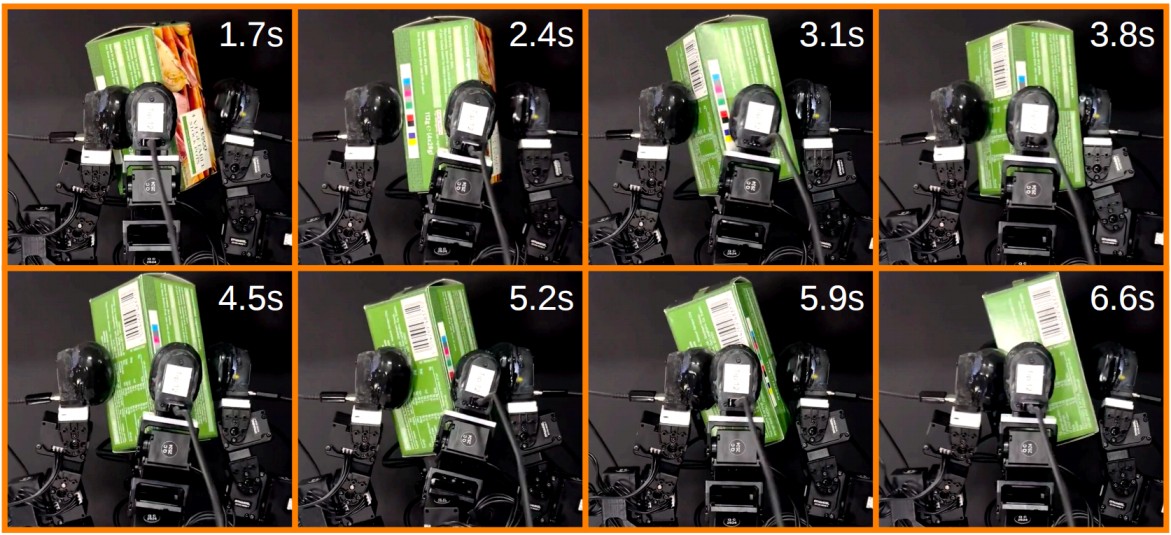} &
        \includegraphics[width=0.49\textwidth]{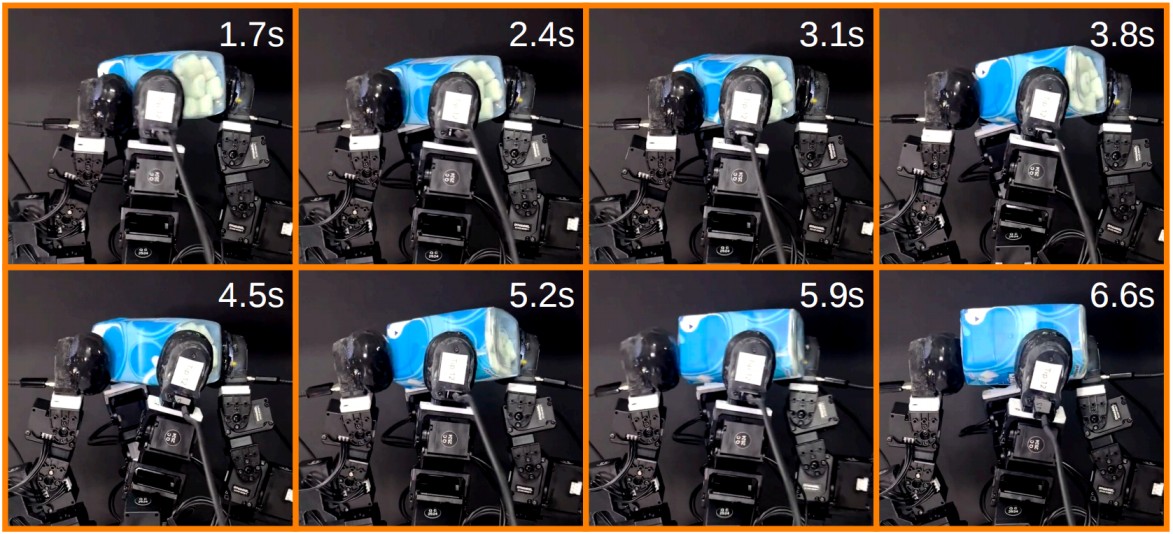}
    \end{tabular}
    \caption{\small Snapshots of real-world in-hand rotation for example objects with different geometries. From top to bottom, the rows show cylindrical objects in cyan-blue frames (toy cylinder and can), round objects in purple frames (peach and apple), and box-shaped objects in orange frames (pot box and gum container).}
    \label{fig:rw_inhand_rotation_snapshots}
    \vspace{1em}
\end{figure*}
The sim-to-real in-hand rotation policies are evaluated on three sets of out-of-distribution (OOD) objects, covering cylindrical, round, and cube-shaped geometries (Fig.~\ref{fig:hand_object_set}). Within each category, the objects vary in size, mass, center of mass, surface friction, texture, and compliance. Each method is evaluated from 10 distinct initial grasp configurations, with the maximum episode duration limited to 30~s. We report the number of completed rotations (Rot.) and the time to termination (TTT). The real-world in-hand manipulation experiments are conducted using a symmetric LEAP Hand equipped with four vision-based tactile fingertips~\cite{he2026tactileinhandobjectrotation}.

The overall results demonstrate the benefit of high-dimensional tactile observations (Table~\ref{table:real_world_rotation_objects}). The policy using contact depth and tactile flow achieves the best rotation performance, completing $1.95 \pm 0.81$ rotations on average, compared with $1.70 \pm 0.78$ for contact depth alone, $1.61 \pm 0.90$ for low-dimensional contact, and $0.82 \pm 0.56$ for proprioception. Both rich-tactile policies also maintain the objects for the full 30-s episode horizon, whereas the LD-contact and proprioception baselines terminate earlier. These results indicate that contact depth substantially improves grasp maintenance, while tactile flow further improves rotation efficiency by capturing contact-induced tangential motion and shear-related deformation cues.

The performance further divides across object-shape categories (Table~\ref{table:real_world_rotation_diff_shapes}). The combined depth-and-flow policy achieves the best performance for all three geometries, obtaining $2.25 \pm 0.07$, $2.43 \pm 0.37$, and $1.33 \pm 1.00$ rotations on cylindrical, round, and cube-shaped objects, respectively. The advantage of tactile flow is most pronounced for cube-shaped objects, for which the performance increases from $1.04 \pm 0.70$ with contact depth alone to $1.33 \pm 1.00$ with depth and flow. This result suggests that tactile flow becomes particularly valuable when the contact geometry changes abruptly and smooth finger--object motion is difficult to maintain.

Notably, all policies in this evaluation are trained in simulation exclusively with capsule-shaped objects using the LEAP Hand. Nevertheless, the rich-tactile policies successfully generalize to real cube-shaped objects, despite the substantial geometric difference. Unlike capsules, which provide smooth and continuously curved surfaces, cubes contain flat faces, sharp edges, and corners that can abruptly change the contact normal and intermittently obstruct finger motion. Contact depth provides detailed local geometric information, while tactile flow captures the evolution of contact-induced deformation and relative surface motion. Together, these observations allow the policy to adapt its finger motions to contact conditions that are not encountered during capsule-only training.

In contrast, AnyRotate~\cite{yang2024anyrotate} represents tactile feedback using only a low-dimensional 2D contact pose and force magnitude. In its original setting, both capsule- and cube-shaped objects were included during simulation training to obtain reliable transfer to real cube-shaped objects. Under our more restrictive capsule-only training protocol, the low-dimensional-contact baseline performs worse than the proprioception-only baseline on cube-shaped objects, achieving $0.28 \pm 0.09$ rotations compared with $0.45 \pm 0.35$. The contact poses and force patterns generated near cube edges and corners can differ substantially from those observed during capsule training, causing the low-dimensional policy to encounter out-of-distribution tactile states. This can result in unstable finger motions, contact jamming, and object drops. These results indicate that low-dimensional contact measurements alone are insufficient for robust geometric generalization, whereas high-dimensional contact depth and tactile flow provide a richer representation of previously unseen contact interactions.

As qualitatively illustrated in Fig.~\ref{fig:rw_inhand_rotation_snapshots}, rotation performance varies substantially with object geometry. The round objects are the easiest to rotate because their shapes are most similar to the capsule-shaped objects used for policy training in simulation. Over the same time interval, the peach and apple rotate by nearly \(180^\circ\), whereas the cylindrical objects rotate by approximately \(120^\circ\) to \(160^\circ\). The box-shaped objects are the most challenging, achieving only about \(45^\circ\) to \(90^\circ\) of rotation over the same interval. Their flat faces and sharp edges interrupt smooth rolling and can obstruct finger motion during rotation. Nevertheless, our rich-tactile policy can continuously rotate these box-shaped objects, whereas the evaluated state-of-the-art baselines fail to complete stable rotation. This result highlights the benefit of high-dimensional contact-depth and tactile-flow observations for generalizing to object geometries that differ substantially from those used during simulation training. Videos showing the corresponding manipulation sequences together with the multimodal tactile streams are provided in the supplementary material.

\begin{table}
    \centering
    \footnotesize
    \setlength{\tabcolsep}{3pt}
    \renewcommand{\arraystretch}{1.05}
    \caption{\small Real-world in-hand rotation performance. Results are reported as mean~$\pm$~standard deviation over evaluation trials.}
    \vspace{-0.5em}
    \label{table:real_world_rotation_objects}
    \begin{tabular}{@{}lcc@{}}
        \toprule
        Method & Rot.~$\uparrow$ & TTT (s)~$\uparrow$ \\
        \midrule
        Prop.~\cite{qi2023inhand}
        & $0.82 \pm 0.56$
        & $27.40 \pm 5.37$ \\

        Prop. + LD Contact~\cite{yang2024anyrotate}
        & $1.61 \pm 0.90$
        & $28.54 \pm 3.57$ \\

        Prop. + Depth (Ours)
        & $1.70 \pm 0.78$
        & $\mathbf{30.00 \pm 0.00}$ \\

        \shortstack[l]{Prop. + Depth + Tactile Flow (Ours)}
        & $\mathbf{1.95 \pm 0.81}$
        & $\mathbf{30.00 \pm 0.00}$ \\
        \bottomrule
    \end{tabular}
    \vspace{1em}
    \centering
    \footnotesize
    \setlength{\tabcolsep}{4pt}
    \renewcommand{\arraystretch}{1.08}
    \caption{\small Real-world in-hand rotation performance (Rot.~$\uparrow$) across different object geometries. Results are reported as mean~$\pm$~standard deviation over evaluation trials.}
    \vspace{-0.5em}
    \label{table:real_world_rotation_diff_shapes}
    \begin{tabular}{@{}lccc@{}}
        \toprule
        & \multicolumn{3}{c}{Object-Shape Category} \\
        \cmidrule(lr){2-4}
        Method
        & Cylinder
        & Round
        & Cube \\
        \midrule

        Prop.~\cite{qi2023inhand}
        & $0.75 \pm 0.35$
        & $1.23 \pm 0.70$
        & $0.45 \pm 0.35$ \\

        Prop. + LD Contact~\cite{yang2024anyrotate}
        & $2.13 \pm 0.53$
        & $1.68 \pm 0.95$
        & $0.28 \pm 0.09$ \\

        Prop. + Depth (Ours)
        & $2.18 \pm 0.11$
        & $2.24 \pm 0.57$
        & $1.04 \pm 0.70$ \\

        \shortstack[l]{Prop. + Depth + Flow (Ours)}
        & $\mathbf{2.25 \pm 0.07}$
        & $\mathbf{2.43 \pm 0.37}$
        & $\mathbf{1.33 \pm 1.00}$ \\

        \bottomrule
    \end{tabular}
\vspace{-1.5em}
\end{table}

\section{Discussion}
\label{sec:discussion}

\subsection{Tactile Flow for Shear-sensitive Manipulation Tasks}
We evaluated simulated tactile flow on a dual-arm object-following task and a substantially more demanding in-hand manipulation task to demonstrate that the proposed representation is structurally necessary for complex contact dynamics.

In the dual-arm object-following task, although tangential interaction is important, the contact geometry remains relatively constrained and low-dimensional. With sufficient engineering effort, it is still possible to design controller rules (e.g., PID controllers) based on explicitly modeled shear quantities or estimated contact states~\cite{lloyd2024shear, freud2025simshear}. In such settings, tactile information can be abstracted into interpretable variables, and the control problem remains close to being fully observable under carefully designed representations.

Dexterous in-hand manipulation, however, fundamentally breaks this paradigm. The contact state becomes distributed, multi-point, and time-varying across a high-DoF hand~\cite{qi2023inhand,yang2024anyrotate,qi2024inhand}, with continual perturbations and subtle object motions. In this regime, explicitly parameterizing contact locations, shear poses, or force vectors becomes combinatorially complex and practically infeasible. More importantly, from a dynamical systems perspective, the task becomes severely partially observable when conditioned on depth-only or instantaneous contact measurements. 

By incorporating dense, high-resolution tactile flow with temporal integration, the observation more closely approximates a sufficient statistic of the contact dynamics, effectively restoring observability in tasks with distributed tangential interaction. Demonstrating success in in-hand manipulation therefore serves as a decisive validation: tactile flow is not simply an auxiliary cue for structured tracking, but a representation that enables learning in regimes where explicit modeling fails and memoryless policies cannot succeed. This establishes tactile flow as a scalable foundation for shear-dependent, high-complexity manipulation beyond what rigid-body depth-based tactile simulation~\cite{gomes2021generation,wang2022tacto,church2022tactile,lin2022tactilegym2} can support.

\subsection{Role of Multimodal Tactile Policy Learning}


The proposed PETS-Net policy architecture combines positional-encoded proprioceptive histories, high-dimensional tactile observations, and distillation latent adaptation. Compared with prior work that uses compact tactile feedback or analytically designed tactile controllers, PETS-Net learns directly from multimodal tactile observations within a teacher--student reinforcement learning framework. The tactile encoder extracts spatial contact features from contact depth and tactile flow, while the adaptation module converts temporal tactile-proprioceptive histories into a latent representation compatible with the privileged teacher policy. This enables efficient learning and deployment of tactile policies for shear-sensitive tasks such as object following, shear recovery, and dexterous in-hand manipulation.

An important benefit of the proposed tactile representation is the complementarity between contact depth and tactile flow. Contact depth provides dense information about where and how deeply contact occurs, while tactile flow provides motion cues about how the contact evolves tangentially. This is particularly useful in tasks where different contact states may have similar instantaneous depth observations but different future contact evolution. Together this allows our policy learning pipeline to learn control directly from the rich multimodal tactile observations generated by TactileLab.

\subsection{Real-to-Sim Tactile Transfer for Sim-to-Real Policy Deployment}

A key design principle of TactileLab is to use the simulated tactile modalities as a common intermediate representation between real tactile sensing and policy learning. Rather than reproducing the full appearance or deformation mechanics of each physical tactile sensor in simulation, contact depth and tactile flow provide physically interpretable representations of normal contact geometry and tangential contact motion. Policies can therefore be trained directly from these representations at scale in simulation, while sensor-specific real-to-sim translators map real tactile observations into the same observation space during deployment. The cross-sensor real-to-sim validation results show that tactile flow can be recovered from tactile sensors with substantially different sensing geometries and surface features, including marker-based TacTip and GelSight sensors as well as 9DTact. The poor performance of the markerless GelSight Mini also highlights an important practical requirement: tangential deformation must remain observable in the real tactile images for reliable tactile-flow recovery.

Together with the real-world dual-arm object-following and dexterous in-hand manipulation experiments, these results demonstrate the complete transfer pathway from real tactile observations to simulated tactile representations and finally to policies trained entirely in simulation and deployed on physical robots. This separation between sensor-specific perception and policy learning reduces the dependence of the control policy on the appearance and sensing mechanism of a particular tactile sensor. In principle, adapting the framework to a new sensor would primarily require learning a new real-to-sim observation mapping into the same contact-depth and tactile-flow space, rather than redesigning the simulated tactile representation or the downstream policy around a sensor-specific tactile signal.

\subsection{Limitations and Future Work}
\label{sec:future_work}

\subsubsection{No-slip Assumption}
In the tasks considered in this work, the real-to-sim tactile transfer assumes that the elastomer remains in sticking contact with the object surface, such that contact deformation is dominated by shear rather than slip. While this assumption is generally satisfied once the policy is sufficiently trained to maintain stable contact, external disturbances or abrupt contact changes may violate this condition, potentially leading to unrecoverable states. Closing the sim-to-real gap on slip would require introducing slip-detection capability, which the sensor used in this work has been demonstrated to be effective at~\cite{jasper2021slip}, although it is an open problem how to include this in tactile simulation.


\subsubsection{Extension to Soft-body Tactile Simulation}

In this work, we investigate simulated tactile flow within a rigid-body simulation framework to obtain an efficient and scalable approximation of shear-relevant contact cues. However, the tactile-flow representation itself is not restricted to rigid-body simulation. The same formulation can be extended to soft-body tactile simulators, where the flow field could be computed from simulated elastomer deformation or rendered tactile image motion. Thus, rigid-body tactile flow provides an efficient abstraction for large-scale policy learning, while soft-body simulators could provide a higher-fidelity source of the same modality when computational cost is less restrictive.





Overall, extending the method to more diverse object shapes will require larger high-dimensional real-to-sim datasets and geometry-aware transfer models. In particular, the model must account for local contact curvature, contact orientation, object shape, and the resulting image-space deformation patterns. This is especially important for tactile flow, since the same tangential contact motion may appear differently depending on the local geometry of the contacting surface. Addressing these challenges could enable a more general tactile interface that maps diverse real sensors and contact geometries into shared physically grounded representations. Such an interface would provide a common basis for learning tactile manipulation policies across sensors, tasks, and robotic embodiments.

\bibliographystyle{unsrt}
\bibliography{manual}

\clearpage

\end{document}